\documentclass[letterpaper, 10 pt, conference]{ieeeconf}

\IEEEoverridecommandlockouts                              

\title{\LARGE \bf
An Iterative LQR Controller for Off-Road and On-Road Vehicles using a Neural Network Dynamics Model
}

\author{Akhil Nagariya$^{1}$ and Srikanth Saripalli$^{2}$
\thanks{$^{1}$Akhil Nagariya is a PhD student in Mechanical Engineering with the Department of Mechanical Engineering, Texas A\&M University,
        College Station, Texas 77845, USA
        {\tt\small akhil.nagariya@gmail.com}}
\thanks{$^{2}$Srikanth Saripalli is with the Faculty of Mechanical Engineering, Department of Mechanical Engineering,
        Texas A\&M University, College Station, Texas 77845, USA
        {\tt\small ssaripalli@tamu.edu}}%
}
\usepackage{amsmath}
\usepackage{algorithmicx}
\usepackage{algorithm}
\usepackage{algpseudocode}
\usepackage{amssymb}
\usepackage{graphicx}
\usepackage{caption}
\usepackage{subcaption}
\usepackage{bm}
\usepackage{float}

\usepackage{etoolbox}
\makeatletter
\AtBeginEnvironment{thebibliography}{%
   \clubpenalty10000
   \@clubpenalty \clubpenalty
   \widowpenalty10000}
\makeatother

\begin{document}

\maketitle	
\thispagestyle{empty}
\pagestyle{empty}

\begin{abstract}
In this work we evaluate Iterative Linear Quadratic Regulator(ILQR) for trajectory tracking of two different kinds of wheeled mobile robots namely Warthog (Fig. \ref{fig:warthog}), an off-road
holonomic robot with skid-steering and Polaris GEM e6  \cite{Gem}, a non-holonomic six seater vehicle (Fig. \ref{fig:golfcart}). We use multilayer neural network to learn the discrete dynamic model of these robots which is
used in ILQR controller to compute the control law. We use model predictive control (MPC) to deal with model imperfections and  perform 
extensive experiments to evaluate the performance of the controller on human driven reference trajectories with 
vehicle speeds of 3m/s-4m/s for warthog and 7m/s-10m/s for the Polaris GEM.

\end{abstract}

\section{INTRODUCTION}
Model based control approaches are used successfully to control complex dynamic systems \cite{pilco}, \cite{minmax}, \cite{legged}, \cite{derl}. These approaches rely on using the dynamic model
of the system to compute the control law for a task in hand. In model based approaches once the controller is developed it can be utilized to perform different types of control
tasks compared to model free approaches where agent has to learn a new policy for every task.
The asymptotic performance of model based approaches is 
generally worse than model free approaches due to inaccuracies in the model \cite{psearch}, \cite{rlsurvey} to deal with this issue researchers often use model based 
controllers in model predictive control (MPC)\cite{IEEEexample:sysid} setting.
Model free approaches require millions of samples 
to learn good policies \cite{tpo}. Collecting samples on a real robot operating in a highly dynamic environments can be
extremely dangerous and renders model free approaches ineffective for these kind of systems. 
Classical model based control on real dynamic systems involve careful system identification \cite{IEEEexample:sysid} that requires 
considerable domain expertise and modeling of the complex dynamics of actuators, tire forces, slip etc in case of wheeled mobile robots. These constraints make the
model free and classical model based control hard and time consuming for real robotic systems. 

In this work we use multilayer neural networks to learn the dynamic model of different types 
of wheeled robot and use ILQR \cite{IEEEexample:gen_syn} as the controller. Neural networks are powerful non-linear function approximators \cite{hornik} \cite{boris} 1\cite{dsurvey} 
and provide an alternative approach for system identification or dynamic modeling of the system  
by only using the data collected from the system. ILQR uses the dynamic model of the
vehicle and provides extra robustness on top of kinematic controllers that cannot deal with the dynamic constraint of the vehicle. In 
section-\ref{rwork} we discuss some of the past research on learning the dynamic model of the system and then discuss some of the model based control approaches which use 
ILQR and are closely related to the work presented in this paper. In section-\ref{approach}  we discuss our approach and in section-\ref{results} present the results of trajectory 
tracking on both Warthog and the Polaris GEM e6 for various reference trajectories . 
\begin{figure}[h]
 \centering{
 \resizebox{75mm}{!}{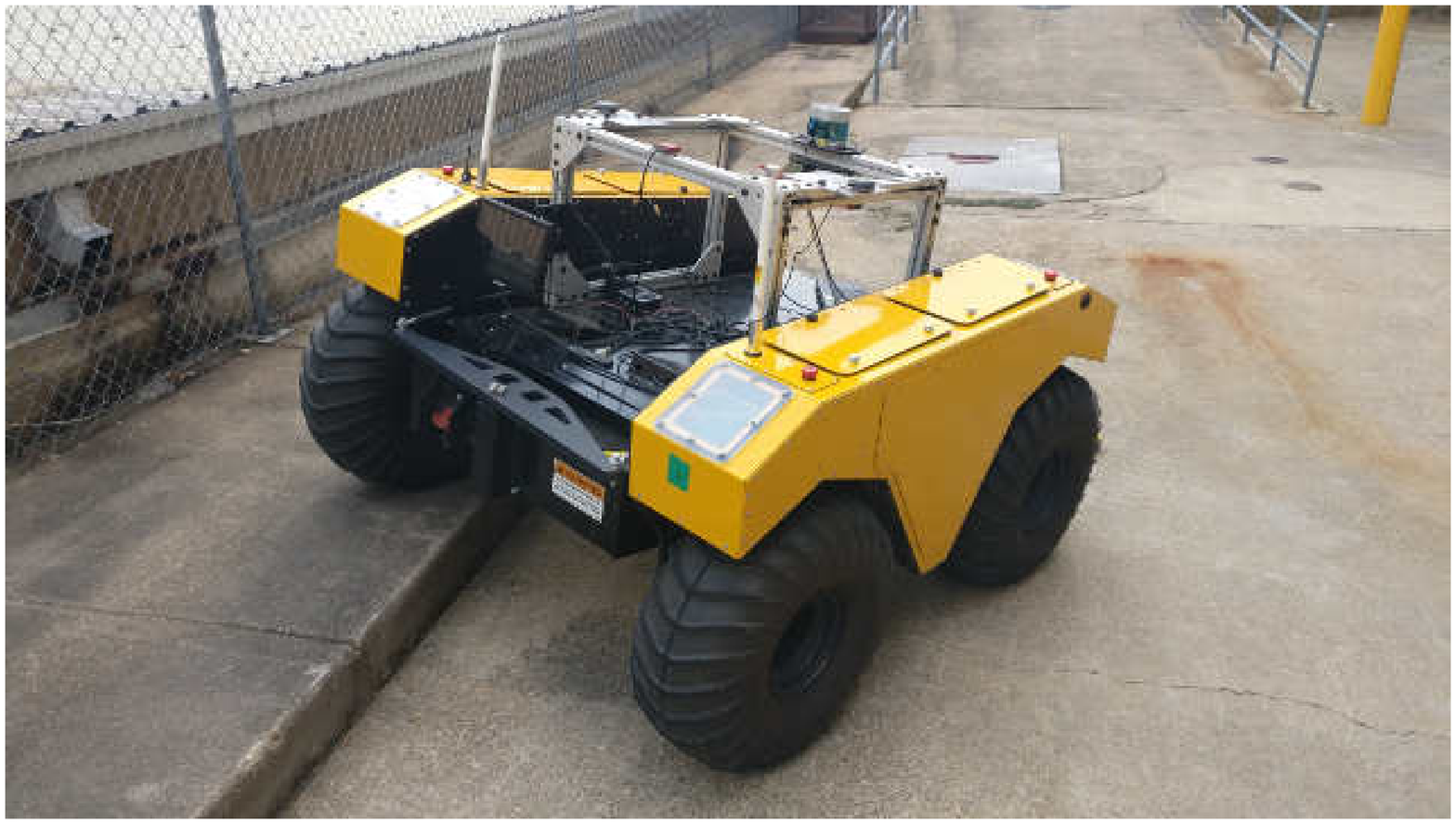}
 \caption{Warthog}
 \label{fig:warthog}
 }
\end{figure}
 \begin{figure}[h]
  \centering{
  \vspace*{1cm}
 \resizebox{75mm}{!}{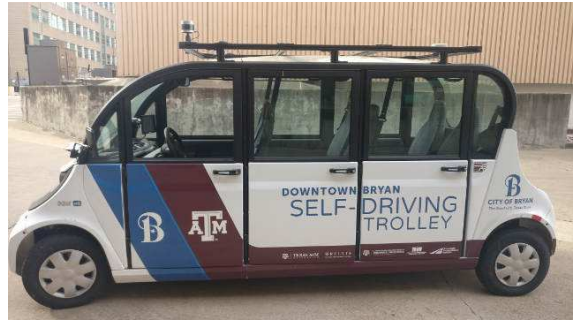}
 \caption{Polaris GEM e6}
 \label{fig:golfcart}
 }
\end{figure}

\section{Related Work}\label{rwork}
In this section we discuss some of the past research on learning non-linear and
stochastic dynamic systems. Then we discuss previous work on using ILQR to control different kinds of robots. 
Finally we discuss some of the past research which combine ILQR and learned model to control a robotic system.    

\subsection{Model Learning}
\cite{IEEEexample:dnn} and \cite{IEEEexample:dnn2}  present first usage of neural networks 
for identification and control of non-linear dynamical systems. \cite{IEEEexample:RBFNN} uses Radial Basis Function Neural Network (RBFNN) to model non-linear stochastic dynamic system. \cite{IEEEexample:MPC} uses neural network to model 
the non-linear dynamics of a neutralization plant and uses this model to control the pH-value. Gaussian processes (GP) are used to model low dimensional stochastic dynamic systems and
preferred over neural networks when only few data-points are available 
\cite{IEEEexample:gpmpc,IEEEexample:gprl2, IEEEexample:localgp, IEEEexample:gpquad, IEEEexample:gpdataeff, IEEEexample:gprldataeff}. 
\cite{IEEEexample:pnn} uses probabilistic neural network to model high dimensional stochastic dynamic systems using significantly fewer samples.
Past research on model learning has focused on learning the dynamic model of manipulators \cite{IEEEexample:gpdataeff}, UAVs \cite{IEEEexample:gpquad}, \cite{IEEEexample:gprl2} or robots in simulation\cite{IEEEexample:gpmpc}, \cite{IEEEexample:gprldataeff}.
In this work we use neural networks to learn the dynamic model of off-road and on-road vehicles and validate the learned model 
by integrating it with a controller for trajectory tracking. 
\subsection{ILQR based controllers}
ILQR is a control analog of the Gauss-Newton method for nonlinear least squares optimization and is a variant of Differential Dynamic Programming (DDP) \cite{IEEEexample:ddprog}. DDP uses
second order derivatives of the dynamic model while ILQR uses first order derivatives to speed up the computation.
ILQR is usually used in an  MPC setting where faster dynamics evaluation is more important than the decrease in performance due to inaccurate 
dynamics approximation \cite{IEEEexample:gen_syn}. 
\cite{IEEEexample:ilqr} shows the first use of ILQR to control non-linear 
biological movement systems.
The authors of \cite{IEEEexample:ilqr} extend their work in \cite{IEEEexample:gen_ilqr} and develop ILQG for constrained nonlinear 
stochastic systems with gaussian noise. \cite{IEEEexample:gen_syn} uses ILQR in MPC settings to deal with model imperfections to control a 22-DoF humanoid in simulation. 
\cite{IEEEexample:ext_ilqr} and \cite{IEEEexample:smooth_ilqr} introduce the concept of ILQR smoothing for non-linear dynamic systems with non-quadratic cost function.
ILQR smoothing converges faster taking only about third of the number of iterations required by other existing ILQR approaches.  
\cite{IEEEexample:control_ddp} introduces control constraint in a DDP \cite{IEEEexample:ddprog} setting, previous approaches enforced the control constraint by clamping on control 
limits, \cite{IEEEexample:control_ddp} demonstrates that the naive clamping methods are inefficient and proposed an algorithm which 
solves a quadratic programming problem subject to box constraints at each time step. \cite{IEEEexample:control_ddp} validates the proposed method 
on three simulated problems including the 36-DoF HRP-2 robot. \cite{IEEEexample:constrained_ilqr} generalizes the control constrained used in \cite{IEEEexample:control_ddp}
and presents an approach that can deal with the complex constraints of the general on-road autonomous driving. \cite{IEEEexample:constrained_ilqr}
validates the approach in simulation for different on road driving scenarios like obstacle avoidance, lane change, car following and on general driving which combines 
all of these different scenarios. 
\section{Model Based Control Using ILQR}\label{approach}
In this section we discuss our approach to learn the dynamic model of both Warthog and the Polaris GEM e6 using multilayer neural networks.
Warthog is an off-road robot capable of climbing hills, moving through dense shrubs, rocky terrain and shallow water bodies with 
maximum speed up to 4.5 m/s. Polaris GEM e6 is a six seater non holonomic vehicle with maximum speed up to 10 m/s. Both vehicles
are equipped with a VectorNav-300 GPS for localization. After we discuss the dynamic modeling of these vehicles we presenth the ILQR controller for Model based control in algorithm-\ref{algo1}.
Finally we define the trajectory tracking problem and discuss our
approach of using ILQR and the learned model to track a reference trajectory.
\subsection{Neural Network based Dynamic Model}\label{nnmodel}

Let $\mathbf{x}_t \in \mathbb{R}^n$ denote the state and $\mathbf{u}_t \in \mathbb{R}^m $ denote the control commands of a system at discrete time instant $t$ (henceforth referred to as time $t$).
The dynamics of the system can be given as follows:
\begin{align}\label{eq:dmodel}
  \mathbf{x}_{t+1} = f(\mathbf{x}_t, \mathbf{u}_t)
\end{align}
For Warthog the state of the system $\mathbf{x}_t \in \mathbb{R}^2$  is given by $(v_t, \omega_t)$ where $v_t$ is
the linear velocity and $\omega_t$ is the angular velocity of the Warthog at time $t$. Control command $\mathbf{u}_t \in \mathbb{R}^2 $ is given by $(v_t^c, \omega_t^c)$ where $v_t^c$ and $\omega_t^c$ 
are the commanded linear and angular velocities respectively at time $t$. The dynamic function $f_w$ for the warthog can now be given as follows:
\begin{align}\label{eq:wmodel}
  \left[\begin{array}{c}
    v_{t+1} \\
    \omega_{t+1}
  \end{array}\right] = f^w\left(\left[\begin{array}{c} v_{t} \\ \omega_t\end{array} \right], \left[\begin{array}{c} v_{t}^c \\ \omega_t^c\end{array} \right]\right) 
\end{align}
For Polaris GEM e6 the state $\mathbf{x}_t \in \mathbb{R}^2$ is given by $(v_t, \dot{\phi}_t)$ where $v_t$ is the linear velocity and $\dot{\phi}_t$ is the steering angle 
rate at time $t$. The control $\mathbf{u}_t \in \mathbb{R}^3$ is given by $(p_t, b_t, \dot{\phi}_t^c)$ where $p_t$ is the throttle, $b_t$ is the
brake and $\dot{\phi}_t^c$ is the commanded steering rate at time $t$. The dynamic function $f_g$ for the Polaris GEM e6 can now be given as:
\begin{align}\label{eq:gmodel}
  \left[\begin{array}{c}
    v_{t+1} \\
    \dot{\phi}_{t+1}
  \end{array}\right] = f^g\left(\left[\begin{array}{c} v_{t} \\ \dot{\phi}_t\end{array} \right], \left[\begin{array}{c} p_{t} \\ b_t \\ \dot{\phi}_t^c\end{array} \right]\right)
\end{align}
We collected the data $(\mathbf{x}_{t+1}, \mathbf{x}_t, \mathbf{u}_t)$ for both Warthog and Polaris GEM e6 by manually driving them using joystick for an hour in on-road and off-road environments. Driving time is decided using trail and error by observing the 
traning and validation losses during training process. The data is sampled 
at 20Hz for the warthog and 30Hz for the Polaris GEM e6 which are fixed hardware specificatons for these platforms.

$\mathbf{x}_t, \mathbf{u}_t$ is used as inputs and $\mathbf{x}_{t+1}$ is used as output to a neural network that learns the dynamic function $f$
by minimizing the mean squared error (MSE) between the predicted output state $\overline{\mathbf{x}}_{t+1}$ and observed output state $\mathbf{x}_{t+1}$.
Two different neural networks are used to learn $f_w$ and $f_g$.
We whiten the data before we feed it to the input layer of the networks. We experimented with multiple architectures and empirically found that
a fully connected neural networks with two hidden layers having 64 units each with ReLU activation function 
performs very well with our controller. 

\subsection{ILQR Controller} \label{controller} 
Consider a non-linear discrete dynamic system:
 \begin{align}\label{eq:model}
 \mathbf{x}_{t+1} = f(\mathbf{x}_t,\mathbf{u}_t)
 \end{align}
 Where $\mathbf{x}_t \in \mathbb{R}^n$ is the state of the system and $\mathbf{u}_t \in \mathbb{R}^m$ is the control input at time $t$.
 The cost $J_i(\mathbf{x}, \mathbf{U}_i)$ represents the cost incurred by the system starting from state $\mathbf{x}$ and following the control $\mathbf{U_i}$ thereafter.
 \begin{align}\label{eq:cost}
 J_i(\mathbf{x}, \mathbf{U}_i)=   \sum_{j=i}^{N-1} l(\mathbf{x}_j, \mathbf{u}_j) + l_f(\mathbf{x}_N) 
 \end{align}
 
 $l(\mathbf{x}_j, \mathbf{u}_j)$ is the cost of executing control $\mathbf{u}_j$ in state $\mathbf{x}_j$ and $l_f(\mathbf{x}_N)$ is the final cost of sate $\mathbf{x}_N$.
We want to find the optimal control $\mathbf{U}_0^{*}(\mathbf{x})$ that minimizes the total cost $J_0(\mathbf{x}, \mathbf{U}_0)$.
\begin{algorithm}[h]
\caption{ILQR algorithm}
\label{algo1}
 \begin{algorithmic}[1]
 \Function{backward\_pass}{$l,l_f, f, T$}
  \State $V_\mathbf{x} \gets l_{f,\mathbf{x}}(\mathbf{x}_n)$
  \State $V_{\mathbf{x},\mathbf{x}} \gets l_{f,\mathbf{x} \mathbf{x}}(\mathbf{x}_n)$
  \State $k \gets [] , K \gets []$
  \For{$i \gets n-1 \textrm{ to } 1$}
    \State $Q_\mathbf{x} \gets l_\mathbf{x}|_{\mathbf{x}_i} + (f_\mathbf{x}^T V_{\mathbf{x}})|_{\mathbf{x}_i}$
    \State $Q_\mathbf{u} \gets l_\mathbf{u}|_{\mathbf{x}_i} + (f_\mathbf{u}^T V_{\mathbf{x}})|_{\mathbf{u}_i,\mathbf{x}_i}$
    \State $Q_{\mathbf{x}\mathbf{x}} \gets l_{\mathbf{x}\mathbf{x}}|_{\mathbf{x}_i} +(f_x^T V_{\mathbf{x}\mathbf{x}}f_x) |_{\mathbf{x}_i} $
    \State $Q_{\mathbf{u}\mathbf{u}}\gets l_{\mathbf{u}\mathbf{u}}|_{\mathbf{u}_i} +(f_u^T V_{\mathbf{x}\mathbf{x}}f_u) |_{\mathbf{u}_i,\mathbf{x}_i,\mathbf{u}_i} $
    \State $Q_{\mathbf{u}\mathbf{x}}\gets l_{\mathbf{u}\mathbf{x}}|_{\mathbf{u}_i,\mathbf{x}_i} +(f_u^T V_{\mathbf{x}\mathbf{x}}f_x) |_{\mathbf{u}_i,\mathbf{x}_i,\mathbf{x}_i} $
    \State $\widetilde{Q}_\mathbf{u} \gets l_\mathbf{u}|_{\mathbf{x}_i} + (f_\mathbf{u}^T (V_{\mathbf{x}} + \mu \mathbf{I}_n))|_{\mathbf{u}_i,\mathbf{x}_i}$
    \State $\widetilde{Q}_{\mathbf{u}\mathbf{u}}\gets l_{\mathbf{u}\mathbf{u}}|_{\mathbf{u}_i} +(f_u^T (V_{\mathbf{x}\mathbf{x}}+\mu \mathbf{I}_n)f_u) |_{\mathbf{u}_i,\mathbf{x}_i,\mathbf{u}_i} $
    \State $\widetilde{Q}_{\mathbf{u}\mathbf{x}}\gets l_{\mathbf{u}\mathbf{x}}|_{\mathbf{u}_i,\mathbf{x}_i} +(f_u^T (V_{\mathbf{x}\mathbf{x}}+\mu \mathbf{I}_n)f_x) 
    _{\mathbf{u}_i,\mathbf{x}_i,\mathbf{x}_i} $
    \State $k[i] \gets -\widetilde{Q}_{\mathbf{u}\mathbf{u}}^{-1}\widetilde{Q}_\mathbf{u}$
    \State $K[i] \gets -\widetilde{Q}_{\mathbf{u}\mathbf{u}}^{-1}\widetilde{Q}_{\mathbf{u}\mathbf{x}}$
    \State $V_\mathbf{x} \gets Q_\mathbf{x} + K^TQ_{\mathbf{u}\mathbf{u}}k + K^TQ_\mathbf{u} + Q_{\mathbf{u}\mathbf{x}}^Tk$
    \State $V_{\mathbf{x}\mathbf{x}} \gets Q_{\mathbf{x}\mathbf{x}}+ K^TQ_{\mathbf{u}\mathbf{u}}K + K^TQ_{\mathbf{u}\mathbf{x}} + Q_{\mathbf{u}\mathbf{x}}^TK$
  \EndFor
  \State \Return{$K, k$}
 \EndFunction
 \Function{Forward\_pass}{$k, K, f, T$}
  \State $\overline{\mathbf{x}}_0 \gets \mathbf{x}_0, U \gets [], X \gets []$
  \For{$i \gets 1 \textrm{ to } N-1} $
  \State $\overline{\mathbf{u}}_i \gets \mathbf{u}_i + \alpha k[i] + K[i](\overline{\mathbf{x}}_i - \mathbf{x}_i)$
  \State $\overline{\mathbf{x}}_{i+1} = f(\overline{\mathbf{x}}_i, \overline{\mathbf{u}}_i)$
  \State $X[i] \gets \overline{\mathbf{x}}_i $
  \State $U[i] \gets \textrm{CLIP}(\overline{\mathbf{u}}_i, \mathbf{u}_{min},  \mathbf{u}_{max})$
  \EndFor
  \State $X[N] \gets \overline{\mathbf{x}}_N$
  \State $T \gets \{X,U\}$
 \State \Return $T$
 \EndFunction
 \Function{ILQR}{$l, l_f, f$}
 \State Sample initial trajectory $T$ using model (\ref{eq:model}) for horizon $N$
 \For{$j \gets 0 \textrm{ to } M$}
    \State $k, K \gets \textrm{ BACKWARD\_PASS}(l, l_f, f, T)$
    \State $T \gets \textrm{ FORWARD\_PASS}(k, K, f, T)$
 \EndFor
 \State \Return $T$ 
 \EndFunction
 \end{algorithmic}

\end{algorithm}
The Pseudo code for ILQR\cite{IEEEexample:ddprog} is given in algorithm-\ref{algo1}. \cite{IEEEexample:ddprog} gives detail discirption
of the algorithm which is avoided here due to space constraints. The parameter $\mu$ is the Levenberg-Merquardt parameter and $\alpha$ is tuned using bactracking line-search. The reader is referred to 
\cite{IEEEexample:gen_syn} for further details on how to tune these parameters.
\subsection{Trajectory Tracking}
In this section we present the development of the trajectory tracking controller for Polaris GEM e6 and 
omit the discussion for warthog due to space constraint but 
similar techniques can be used to develop a trajectory tracking controller for warthog as well. Let 
$\mathbf{s}_i = \{x_i, y_i, \theta_i, \phi_i, v_i, \dot{\phi}_i\}$ represent the state of the
Polaris GEM e6 with wheel base distance $L$, at discrete instant $i$  where $\{x_i, y_i, \theta_i\}$ is the pose, $\phi_i$ is the steering angle, $v_i$ is the velocity and $\dot{\phi}_i$ is the
rate of change of steering angle at discrete instant $i$. The control command is given by $\mathbf{u}_i = \{a_i, b_i, \dot{\phi}^c_i\}$, here 
$a_i$ is the pedal input, $b_i$ is the brake input and $\dot{\phi}^c_i$ is the commanded rate of change of steering angle at discrete instant $i$.
We represent the state transition function with $\pi$: 
\begin{align}\label{statet}
  \mathbf{s}_{i+1} = \pi(\mathbf{s}_i, \mathbf{u}_i)
\end{align}
Using the dynamic functions $\{f_v^g, f_{\dot{\phi}}^g\}$ given in 
section \ref{nnmodel} and following the bicycle model, $\pi$ can be defined by following equations:
\begin{align}\label{stateteq}
  x_{i+1} &= x_i + v_i cos(\theta_i)\Delta t \nonumber \\
  y_{i+1} &= y_i + v_i sin(\theta_i)\Delta t \nonumber \\
  \dot{\phi}_{i+1} &= f^g_{\dot{\phi}}\left(\left[\begin{array}{c} v_{i} \\ \dot{\phi}_i\end{array} \right], \left[\begin{array}{c} a_{i} \\ b_i \\ \dot{\phi}_i^c\end{array} \right]\right) \nonumber \\
  \phi_{i+1}& = \phi_i + \dot{\phi}_i \Delta t \nonumber \nonumber \\
  \theta_{i+1} & = \theta_i + \frac{v_i tan(\phi_i)}{L} \nonumber \\ 
  v_{i+1} &= f^g_v\left(\left[\begin{array}{c} v_{i} \\ \dot{\phi}_i\end{array} \right], \left[\begin{array}{c} a_{i} \\ b_i \\ \dot{\phi}_i^c\end{array} \right]\right)  
\end{align}
\begin{figure}[h]
 \centering{
 \resizebox{75mm}{!}{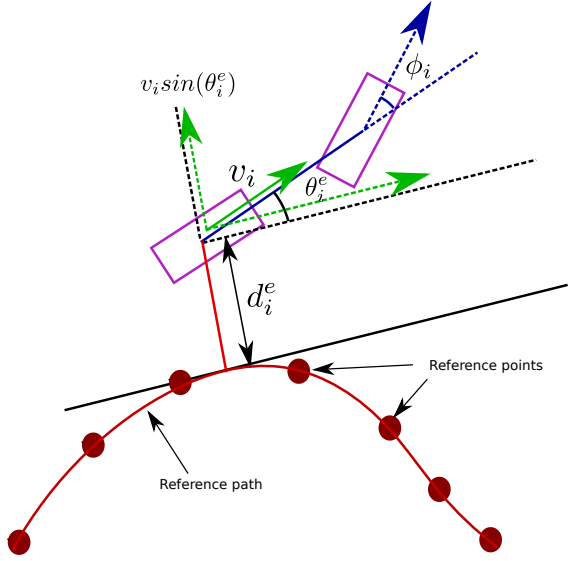}
 \caption{Figure shows bicycle mode of Polaris GEM e6 and error state w.r.t a reference trajectory}
 \label{fig:topview}
 }
\end{figure}

Given a set of $M$ ordered poses with velocities, we 
fit a cubic spline to them and obtain a reference trajectory. `Fig. \ref{fig:topview}' shows bicycle 
model of the Polaris GEM e6 in a typical state $\mathbf{s}_i$, the pink rectangles represent the two wheels, maroon circles 
represent the reference points and the red curve represents the fitted cubic spline. 
For every state $\mathbf{s}_i$ we define an error state $\bm{\psi}_i$ with respect to this reference trajectory  
as a 9-tuple $\{d^e_i, \theta^e_i, v^e_i, \dot{d}^e_i, \dot{\theta}^e_i, \dot{v}^e_i, v_i, \dot{\phi}_i, \phi_i \}$. As shown in  
`Fig. \ref{fig:topview}', $d^e_i$ is the perpendicular distance of a robot
in state $\mathbf{s}_i$ from the reference trajectory, $\theta^e_i$ is the heading error of the robot w.r.t the reference trajectory, $v^e_i$ is the velocity error corresponding to the closest
point on the reference trajectory ($v^e_i = v_i - v_p$, here $v_p$ is the velocity of the closest point on the reference trajectory),  $v_i, \dot{\phi}_i$ and $\phi_i$ are copied from the state $\mathbf{s}_i$.
We use error state $\bm{\psi}_i$ for the ILQR states which encodes all the errors from the reference trajectory. 
Given the error state $\bm{\psi}_i$ and control $u_i$ at a discrete instant $i$ the next error state $\bm{\psi}_{i+1}$ is given:
\begin{align}\label{eq:gamma}
  \bm{\psi}_{i+1} =\gamma(\bm{\psi}_i, \mathbf{u}_i)  
\end{align}
$\gamma$ can be defined by following equations:
\begin{align}\label{eq:gamma}
  d^e_{i+1}& = d^e_i + \dot{d}^e_i \Delta t \nonumber \\
  \theta^e_{i+1} &= \theta^e_i + \dot{\theta}^e_i \Delta t \nonumber \\
  v^e_{i+1} &= v^e_i + \dot{v}^e_i \Delta t \nonumber \\
  \dot{d}^e_{i+1} &= (v^e_i + \dot{v}^e_i \Delta t + v^p_i) sin (\theta^e_i + \dot{\theta}^e_i \Delta t) \nonumber \\
  \dot{\theta}^e_{i+1} &= \frac{(v^e_i + \dot{v}^e_i \Delta t + v^p_i) tan (\phi_i + \dot{\phi}_i \Delta t)}{L} \nonumber \\
  \dot{v}^e_{i+1} &= (v_{i+1} - v_{i})/\Delta t \nonumber \\
  v_{i+1} &= f^g_v\left(\left[\begin{array}{c} v_{i} \\ \dot{\phi}_i\end{array} \right], \left[\begin{array}{c} a_{i} \\ b_i \\ \dot{\phi}_i^c\end{array} \right]\right)  \nonumber \\
  \dot{\phi}_{i+1} &= f^g_{\dot{\phi}}\left(\left[\begin{array}{c} v_{i} \\ \dot{\phi}_i\end{array} \right], \left[\begin{array}{c} a_{i} \\ b_i \\ \dot{\phi}_i^c\end{array} \right]\right)  \nonumber \\
  \phi_{i+1} &= \phi_i + \dot{\phi}_i \Delta t 
\end{align}
The cost $l(\bm{\psi}_i, \mathbf{u}_i)$ of executing $\mathbf{u}_i$ in error state $\bm{\psi}_i$ is given as follows:
\begin{align}\label{eq:lcost}
  l(\bm{\psi}_i, \mathbf{u}_i) = \bm{\psi}_i^T A \bm{\psi}_i +\bm{u}_i^T B \bm{u}_i 
\end{align}
Here $A$ and $B$ are diagonal weight matrices with last 3 diagonal elements of $A$ equal to zero since 
we only care about driving the error terms in (\ref{eq:gamma}) to zero.
 For a state $\bm{\psi}$ the final cost $l_f(\bm{\psi})$ is given as follows:
\begin{align}\label{eq:lfcost}
  l_f(\bm{\psi}) = \bm{\psi}^T A \bm{\psi}
\end{align}
We can now define the trajectory tracking problem for our vehicle with a given 
reference trajectory as finding the optimal control sequence $\{\mathbf{u}_0, \mathbf{u}_1, ..., \mathbf{u}_{N-2}\}$ for horizon $N$ 
that minimizes the following cost:
\begin{align}\label{eq:fincost}
  \sum_{i=0}^{N-2}l(\bm{\psi}_i, \mathbf{u}_i)  + \bm{\psi}_{N-1}^T A \bm{\psi}_{N-1}
\end{align}
Subject to the constraints:
\begin{align}\label{eq:gammconst} 
  \bm{\psi}_{i+1} =\gamma(\bm{\psi}_i, \mathbf{u}_i) \quad \forall i \in \{0, 1, ..., N-2\} 
\end{align}

(\ref{eq:fincost}) and (\ref{eq:gammconst}) transform the trajectory tracking problem to a standard 
ILQR problem defined in section \ref{controller}, and hence can be solved by algorithm \ref{algo1}. 
\section{Results and future work}\label{results}
\begin{figure}[H]
\onecolumn
\begin{tabular}{ccccc}
\centering
 \subcaptionbox{\label{f1:traj}}{\includegraphics[scale=0.175]{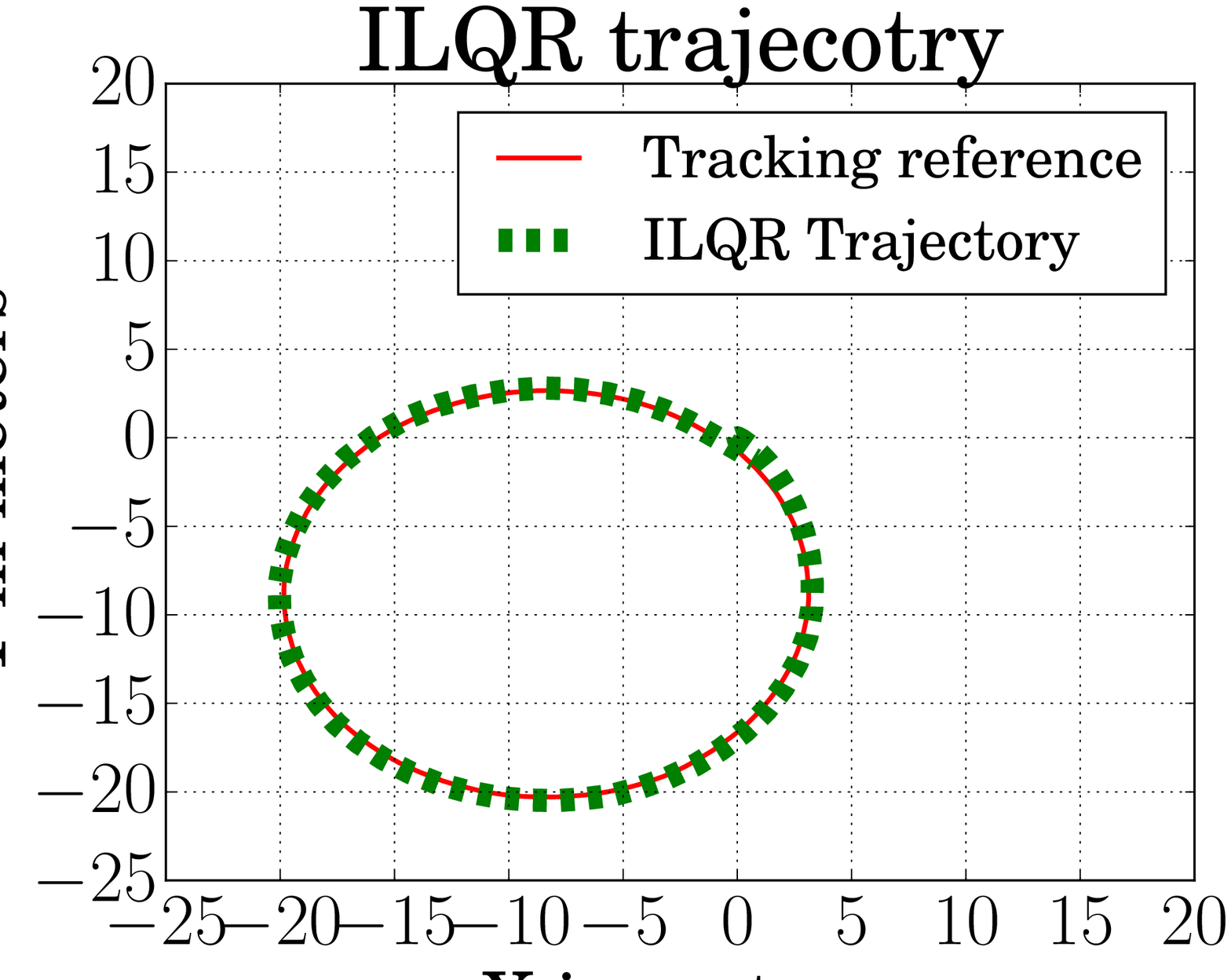}} &
 \subcaptionbox{\label{f1:vel}}{\includegraphics[scale=0.175]{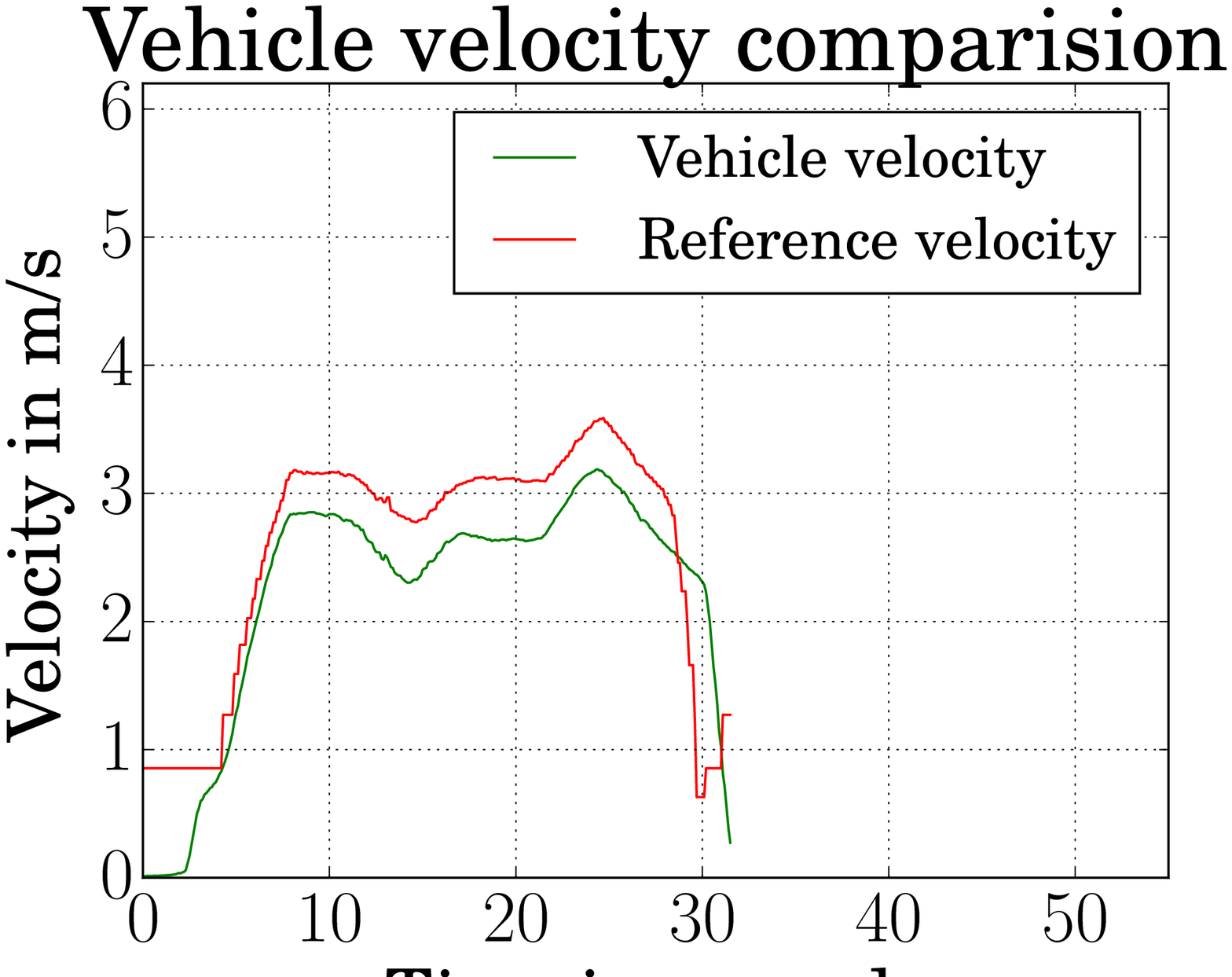}} &
 \subcaptionbox{\label{f1:pedal}}{\includegraphics[scale=0.175]{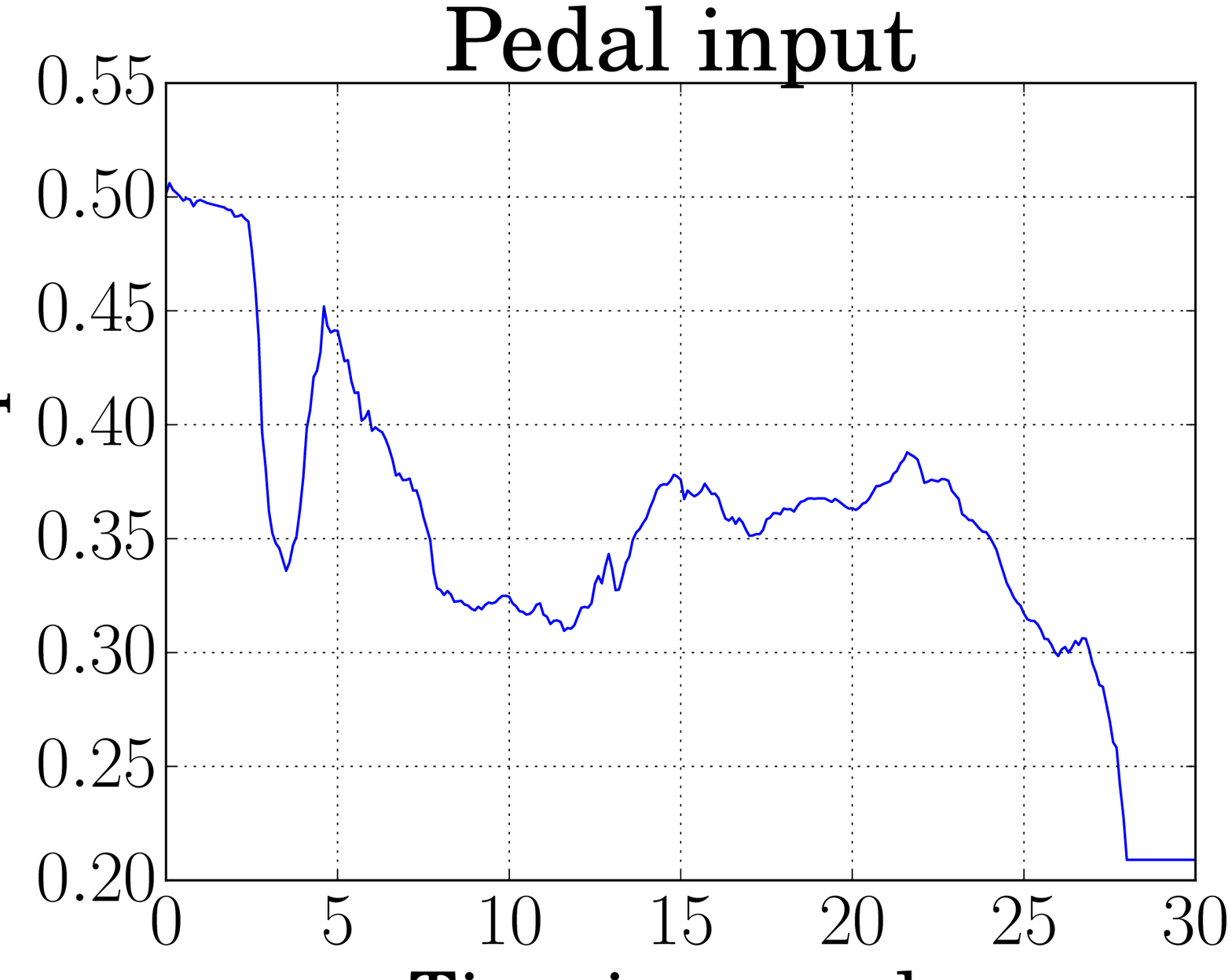}} &
 \subcaptionbox{\label{f1:steer}}{\includegraphics[scale=0.175]{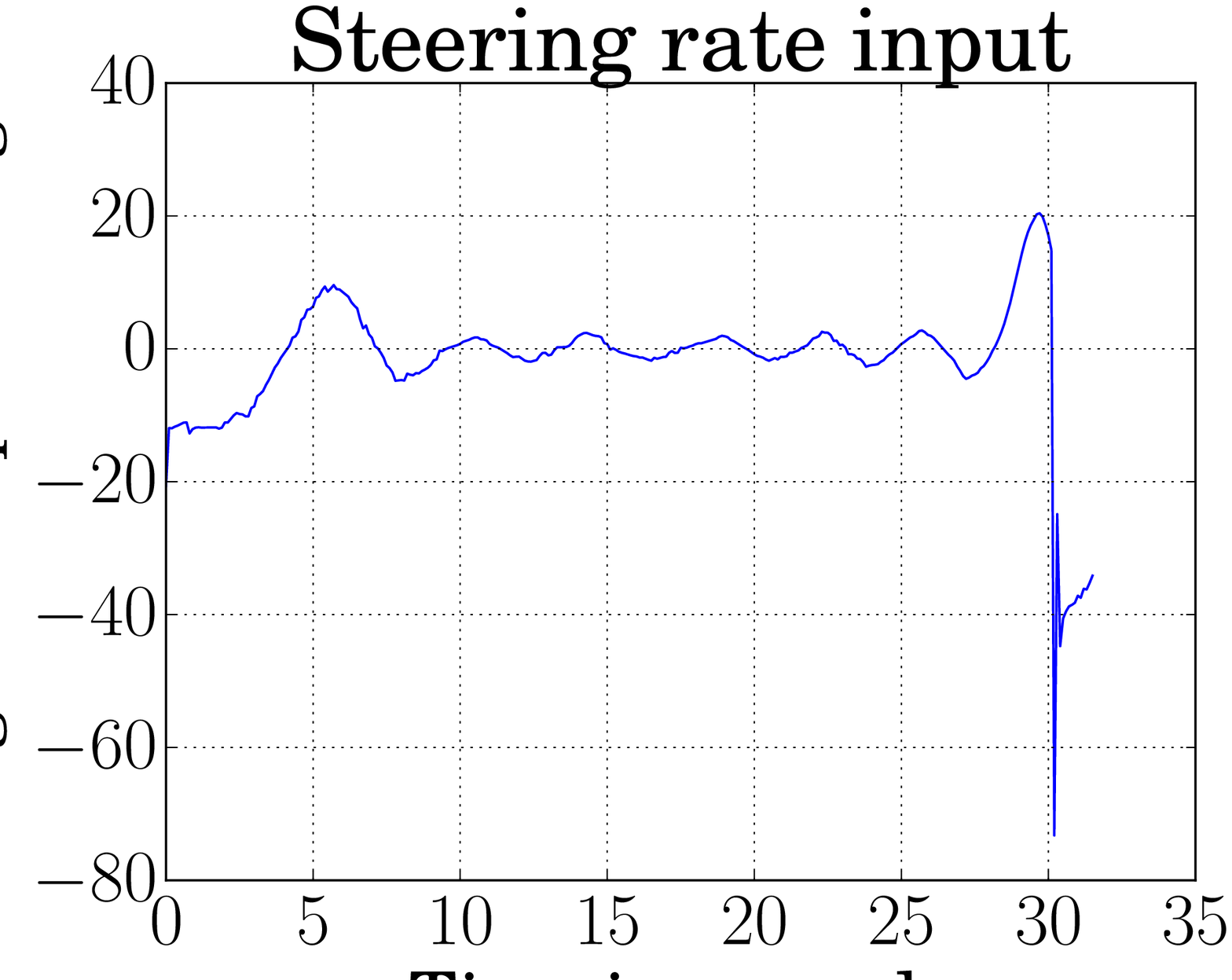}} &
 \subcaptionbox{\label{f1:brake}}{\includegraphics[scale=0.175]{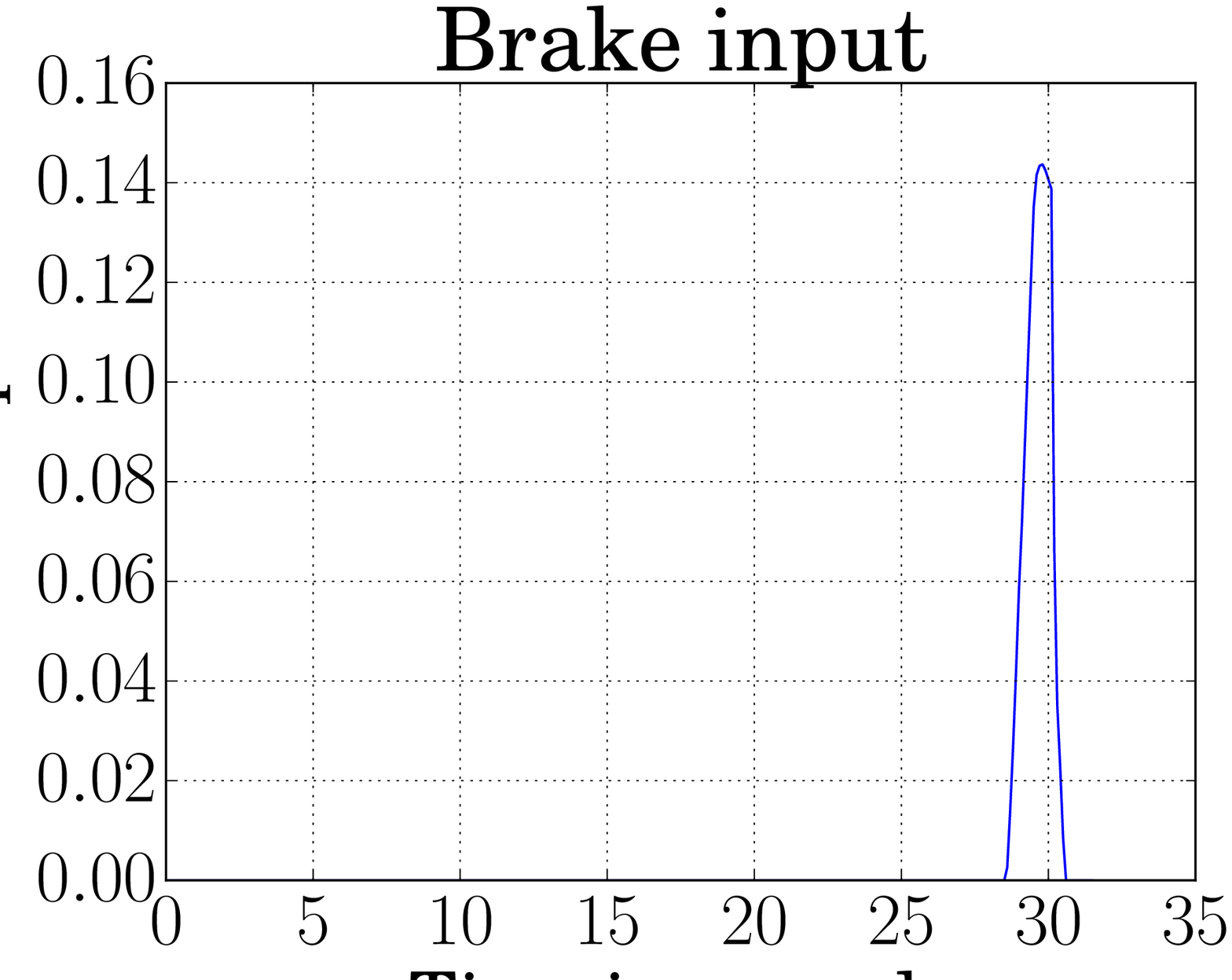}} \\
\end{tabular}
\caption{Polaris GEM e6 Circular trajectory response}
\label{circle}
\end{figure}

\begin{figure}[H]
\onecolumn
\begin{tabular}{ccccc}
\centering
 \subcaptionbox{\label{f2:traj}}{\includegraphics[scale=0.175]{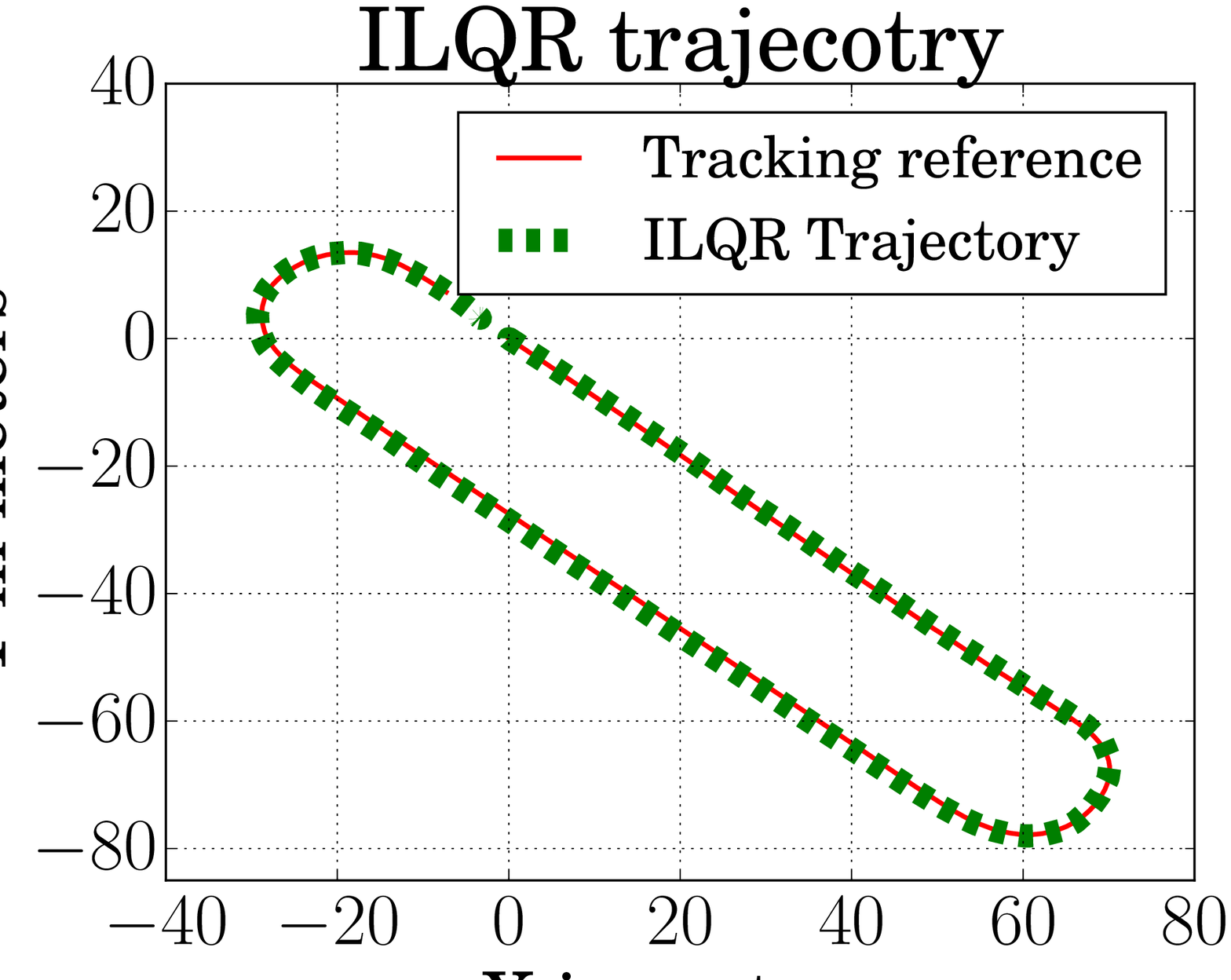}} &
 \subcaptionbox{\label{f2:vel}}{\includegraphics[scale=0.175]{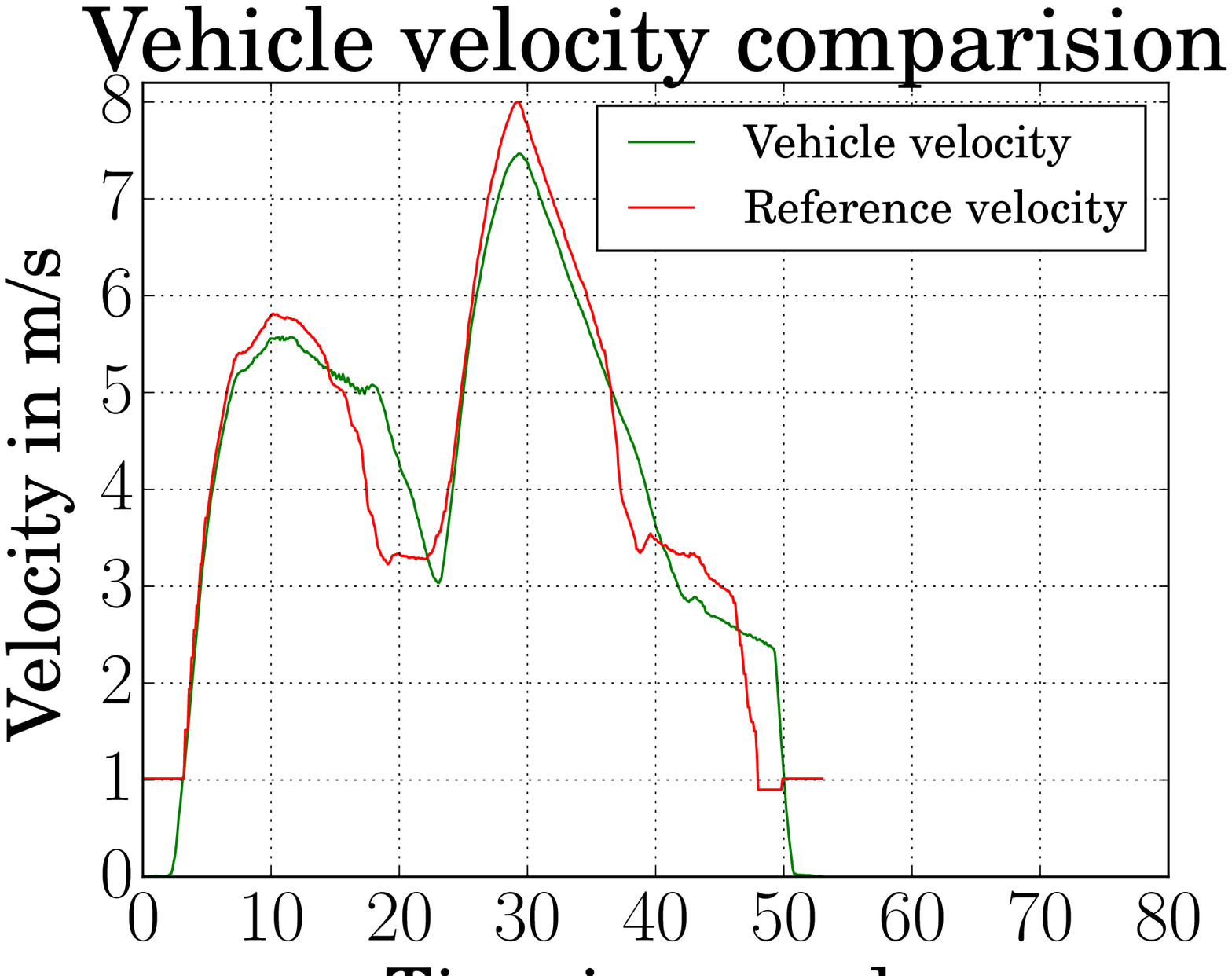}} &
 \subcaptionbox{\label{f2:pedal}}{\includegraphics[scale=0.175]{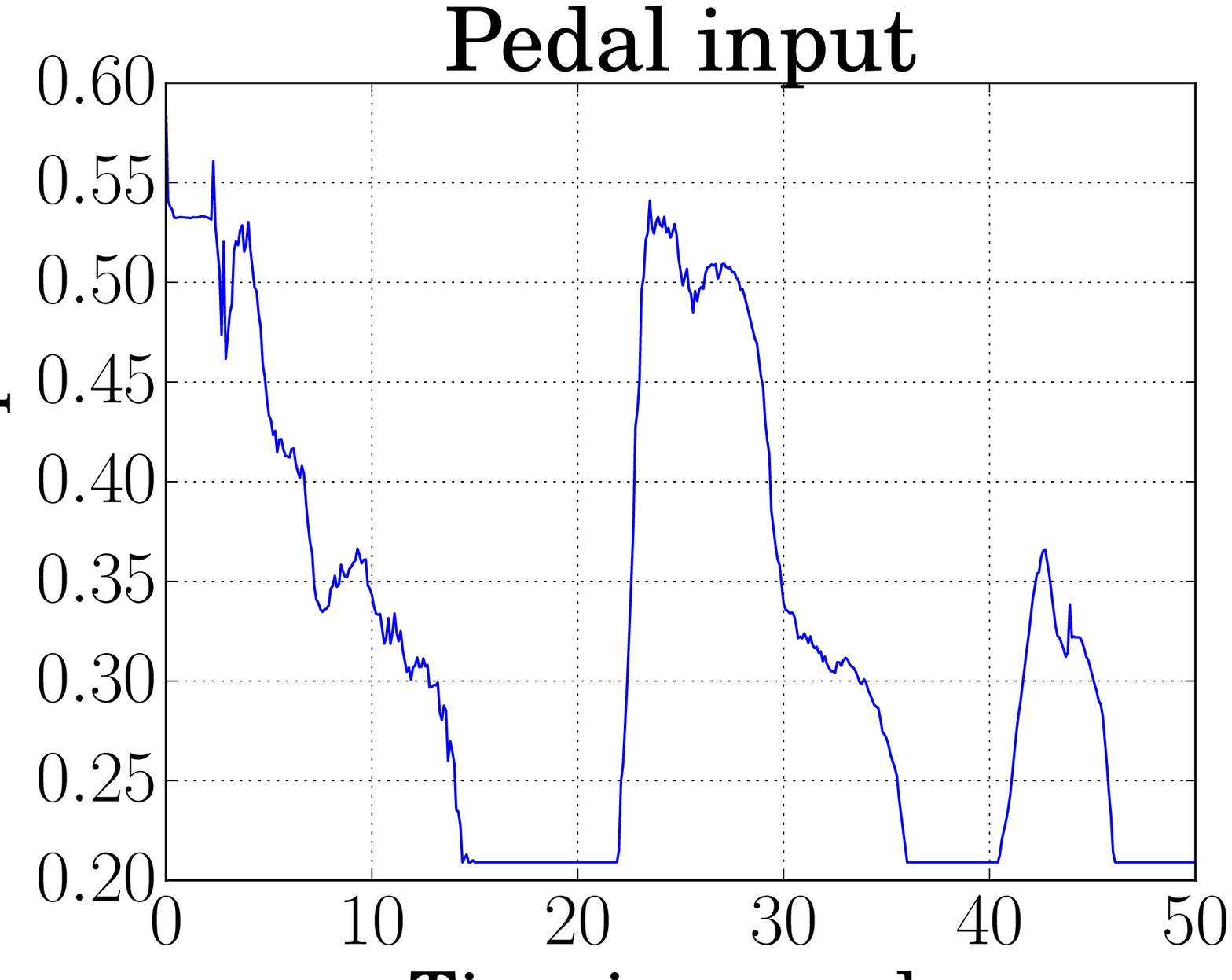}} &
 \subcaptionbox{\label{f2:steer}}{\includegraphics[scale=0.175]{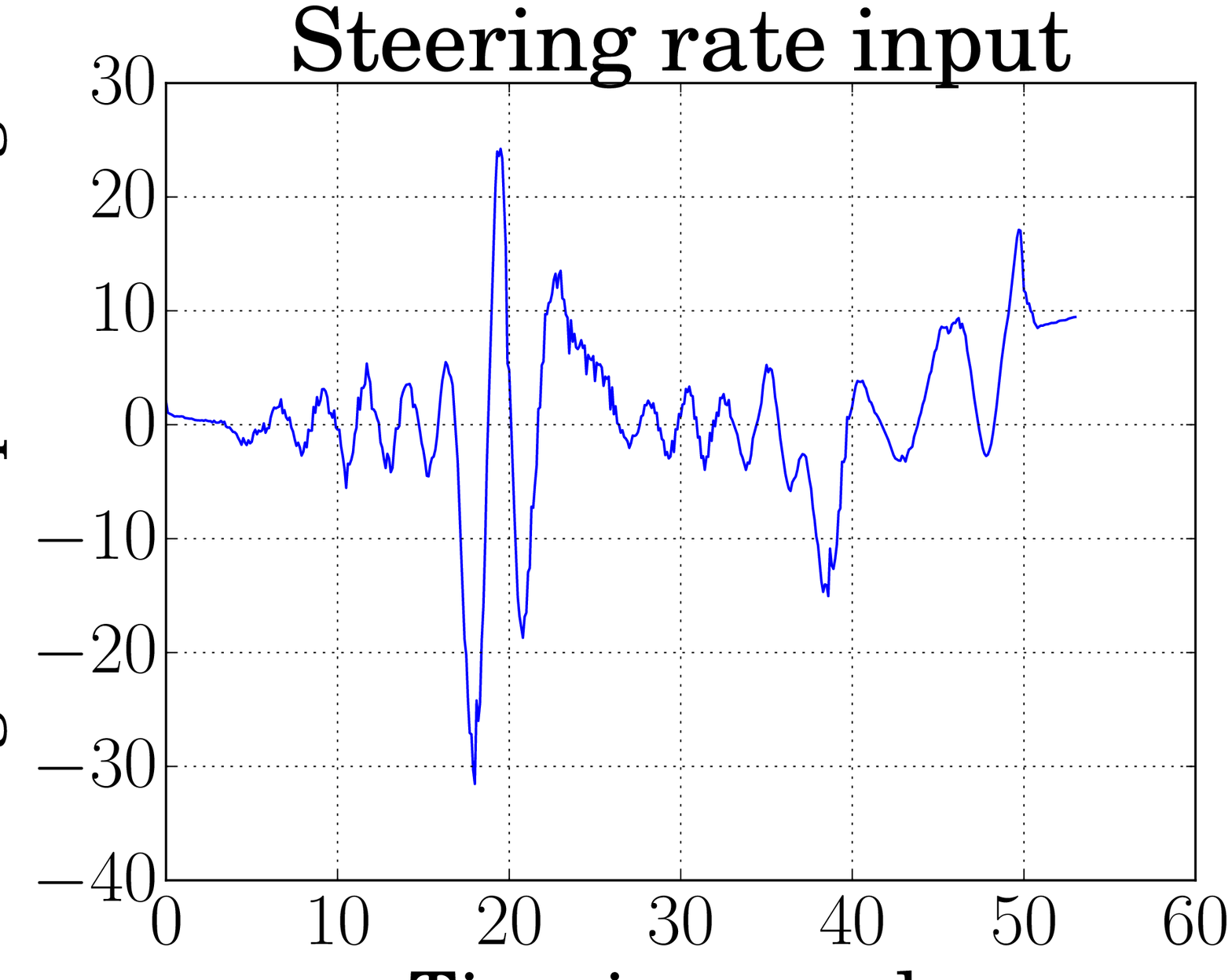}} &
 \subcaptionbox{\label{f2:brake}}{\includegraphics[scale=0.175]{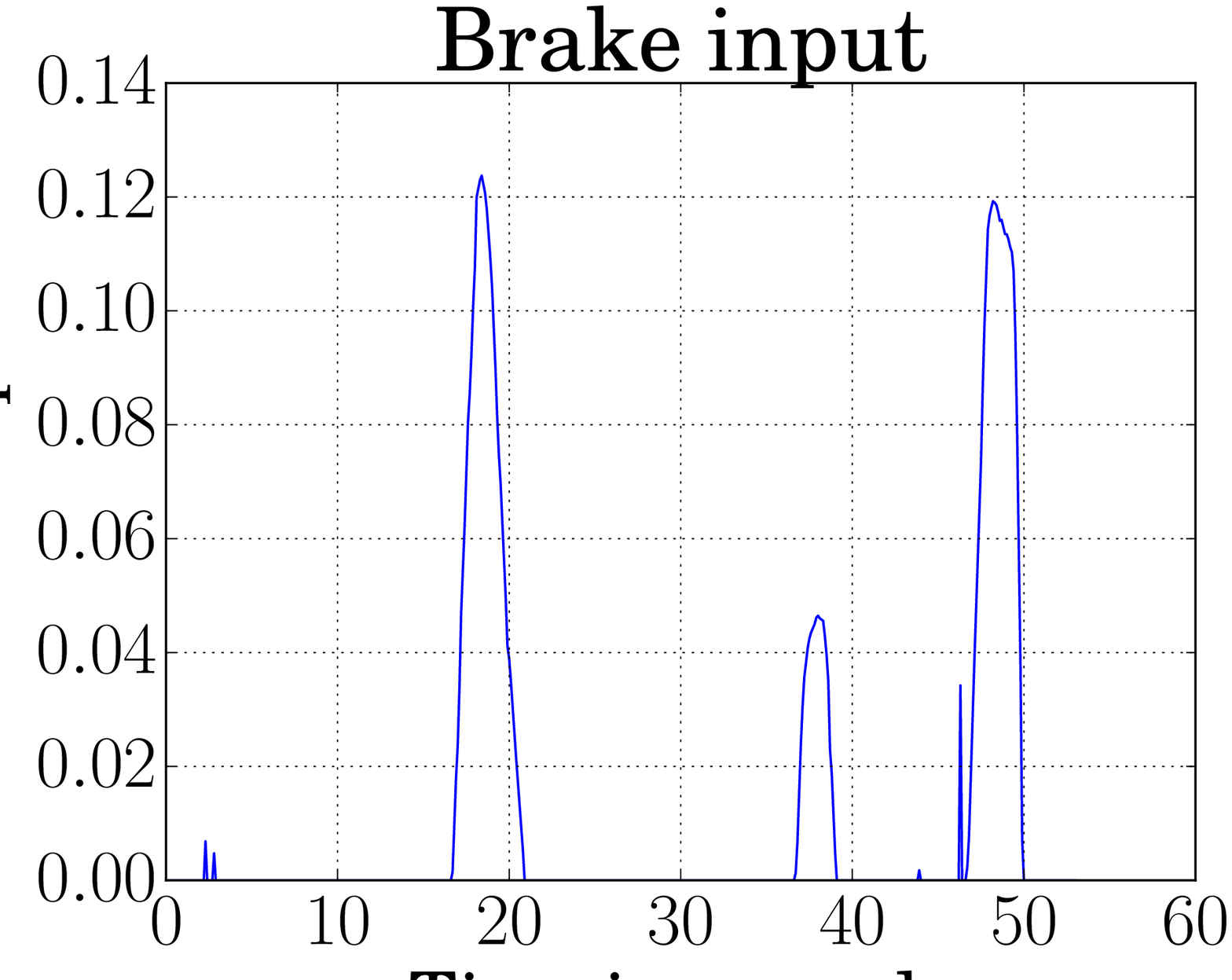}} \\
\end{tabular}
\caption{Polaris GEM e6 Oval trajectory response}

\label{oval}
\end{figure}

\begin{figure}[H]
\onecolumn
\begin{tabular}{ccccc}
\centering
 \subcaptionbox{\label{f3:traj}}{\includegraphics[scale=0.175]{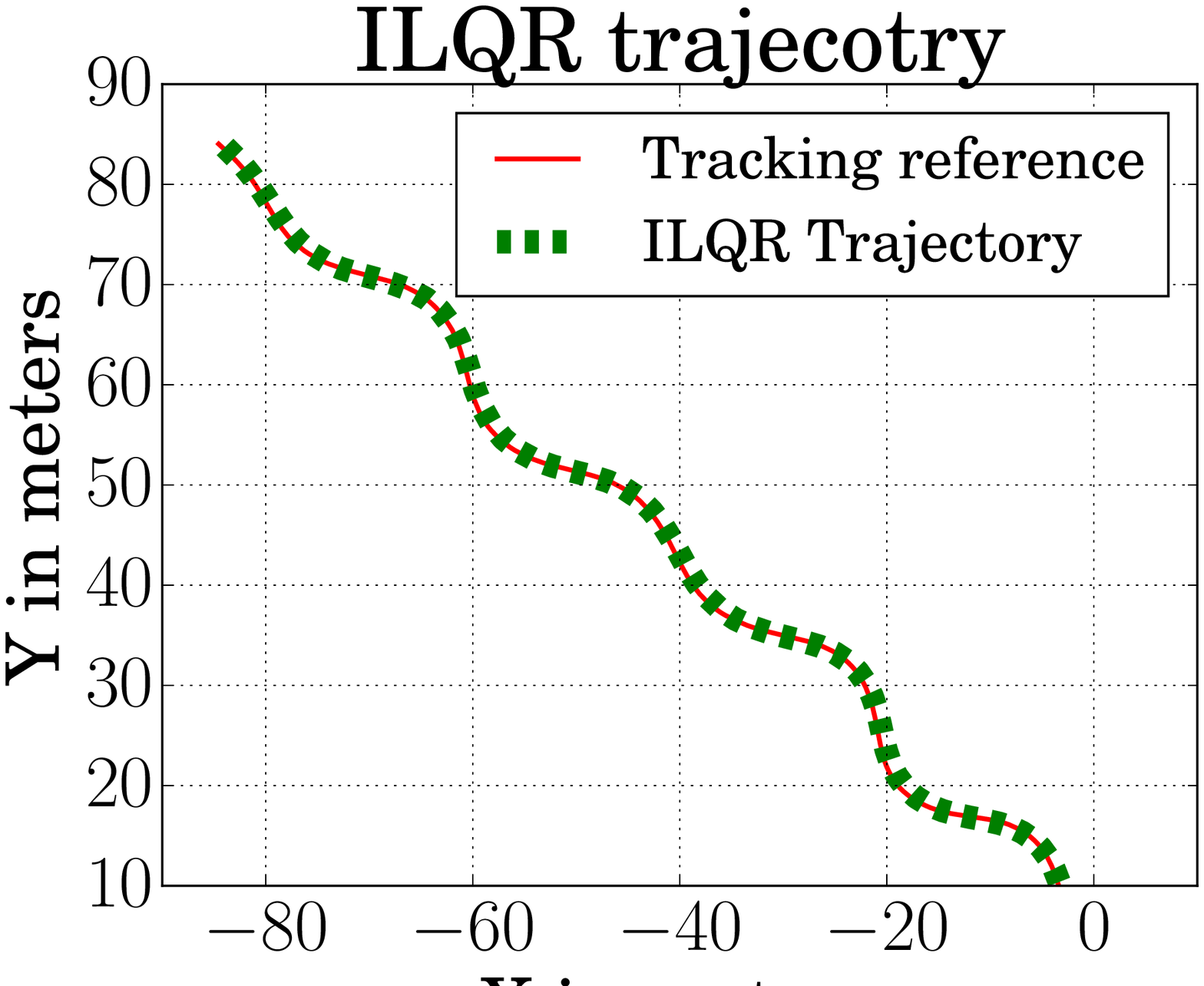}} &
 \subcaptionbox{\label{f3:vel}}{\includegraphics[scale=0.175]{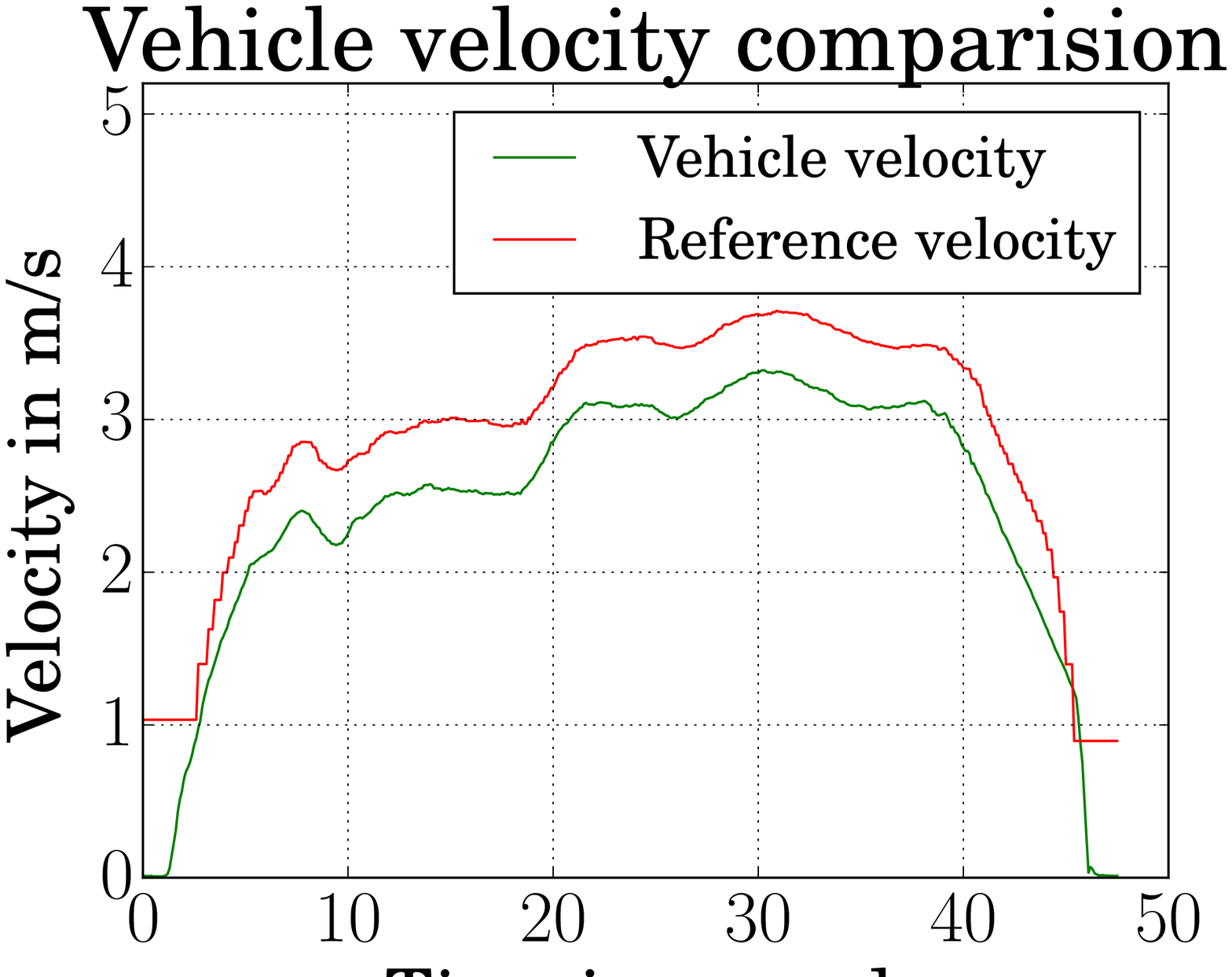}} &
 \subcaptionbox{\label{f3:pedal}}{\includegraphics[scale=0.175]{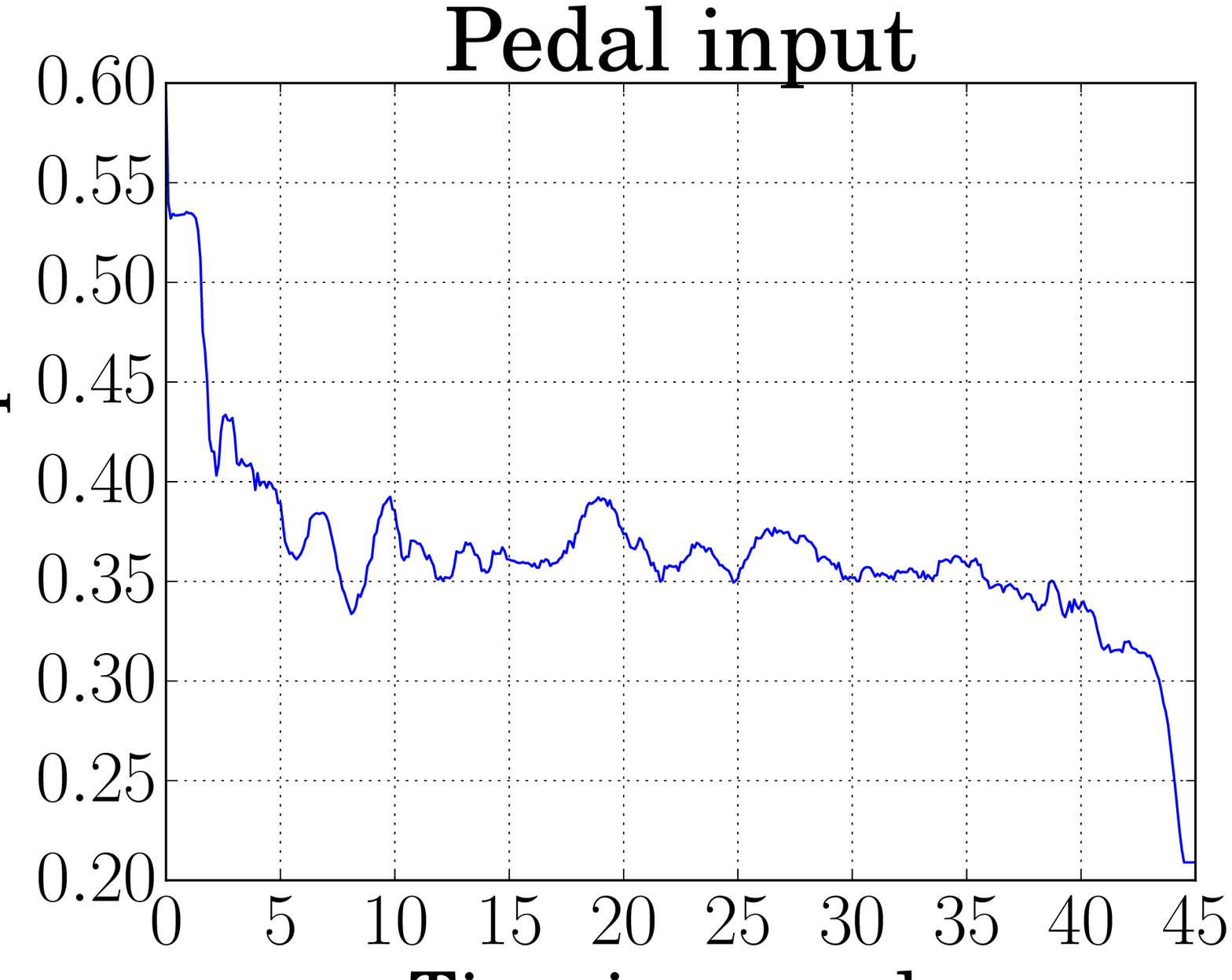}} &
 \subcaptionbox{\label{f3:steer}}{\includegraphics[scale=0.175]{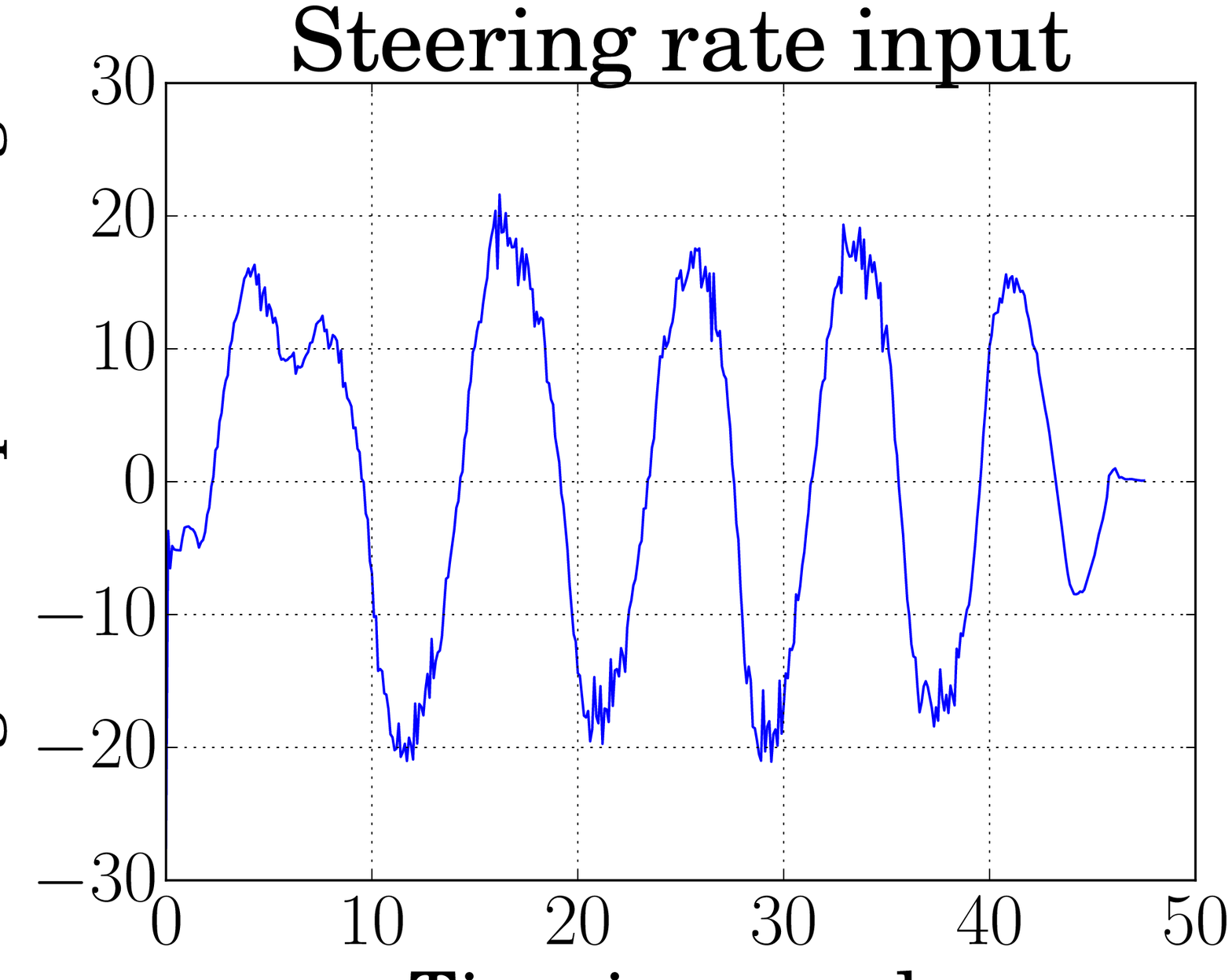}} &
 \subcaptionbox{\label{f3:brake}}{\includegraphics[scale=0.175]{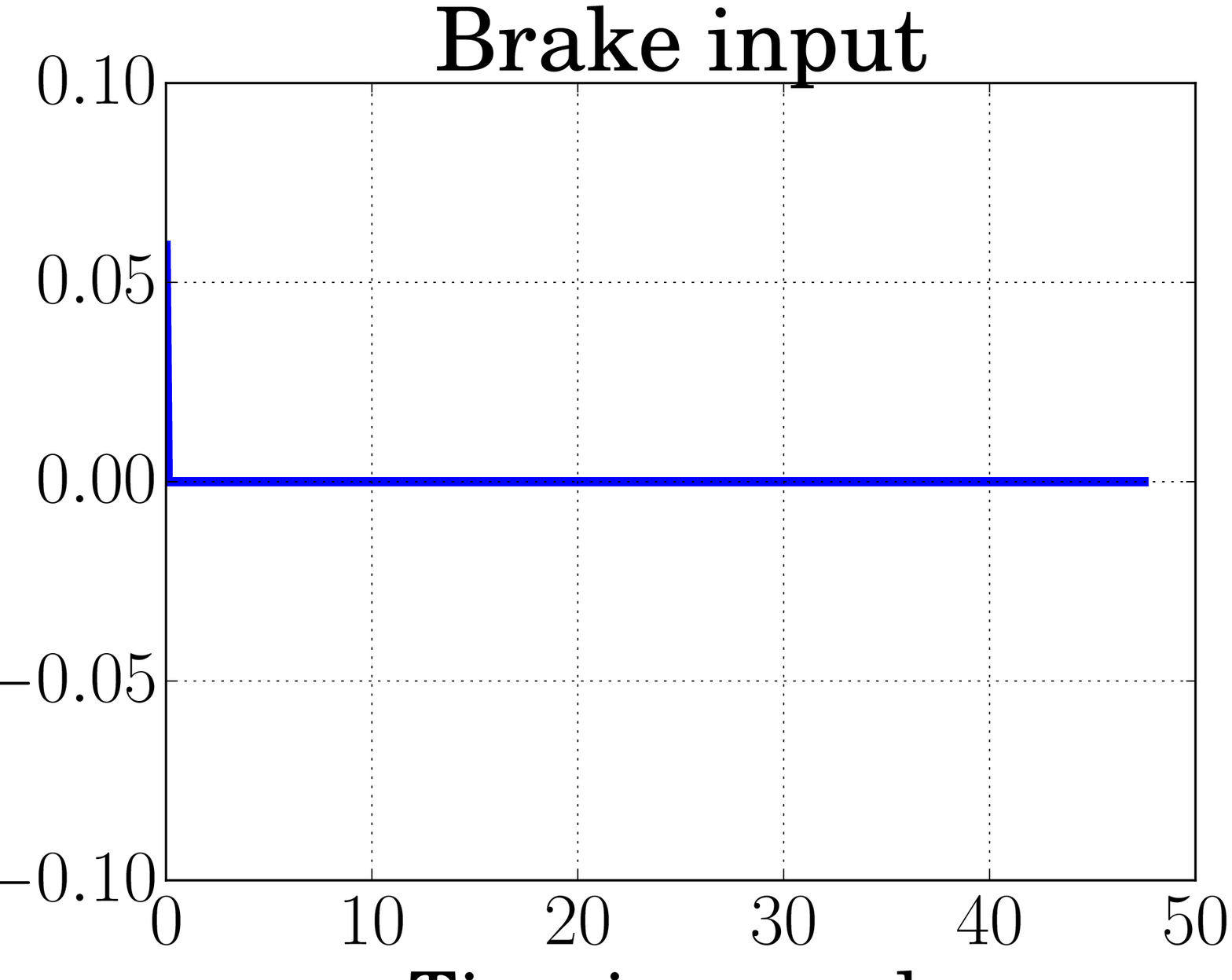}} \\
\end{tabular}
\caption{Polaris GEM e6 Snake trajectory response}
\label{snake}
\end{figure}

\begin{figure}[H]
\onecolumn
\begin{tabular}{ccccc}
\centering
 \subcaptionbox{\label{f4:traj}}{\includegraphics[scale=0.175]{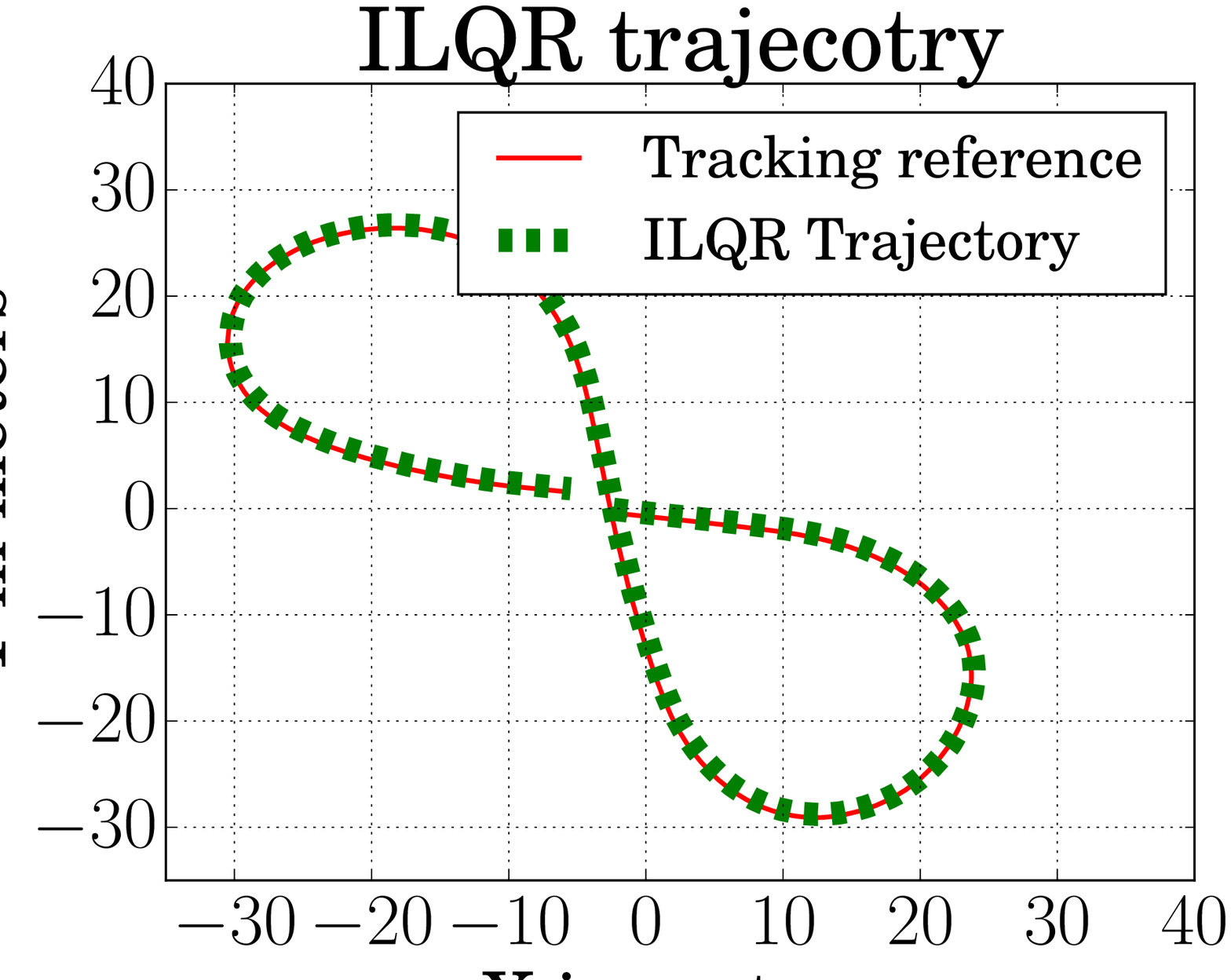}} &
 \subcaptionbox{\label{f4:vel}}{\includegraphics[scale=0.175]{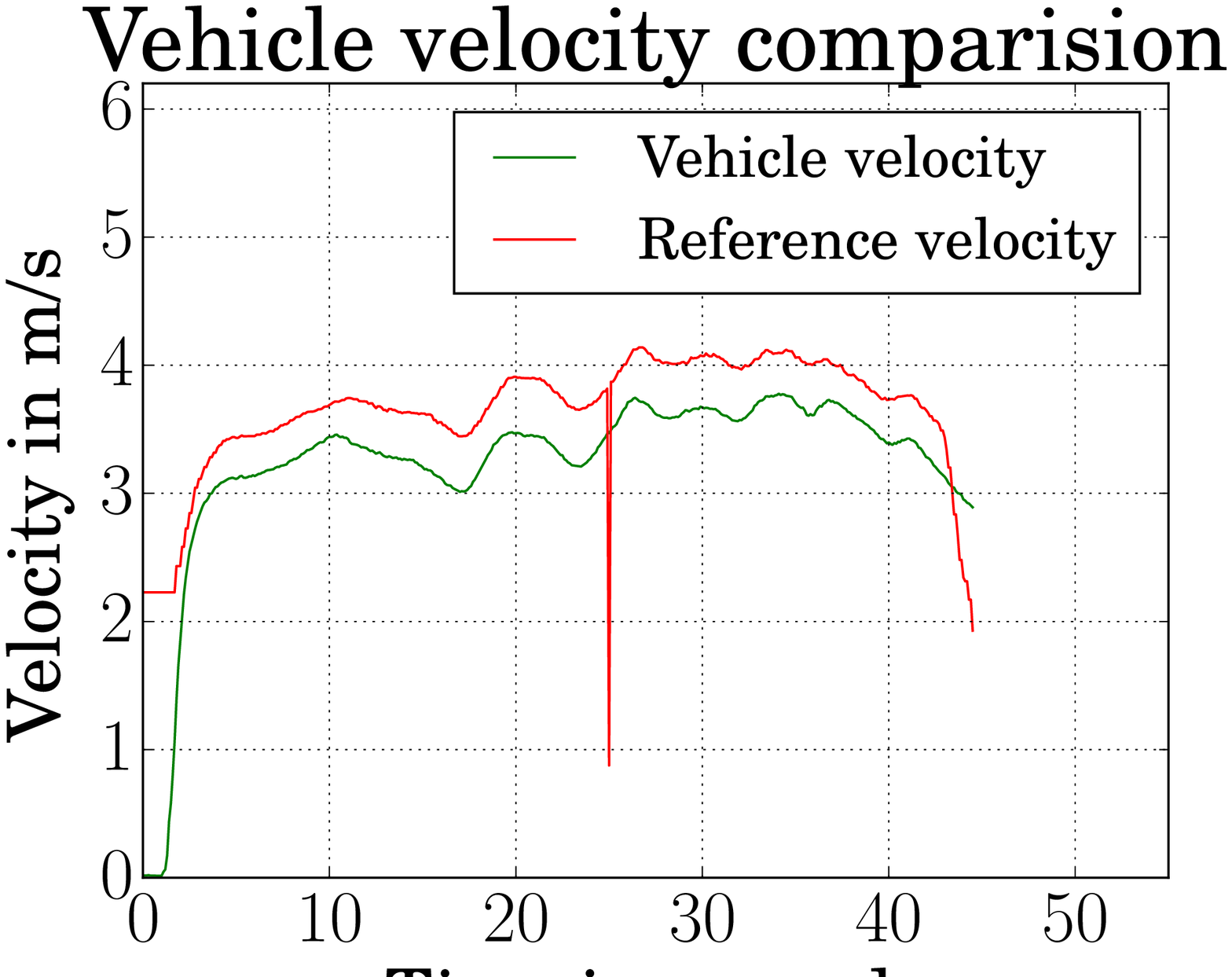}} &
 \subcaptionbox{\label{f4:pedal}}{\includegraphics[scale=0.175]{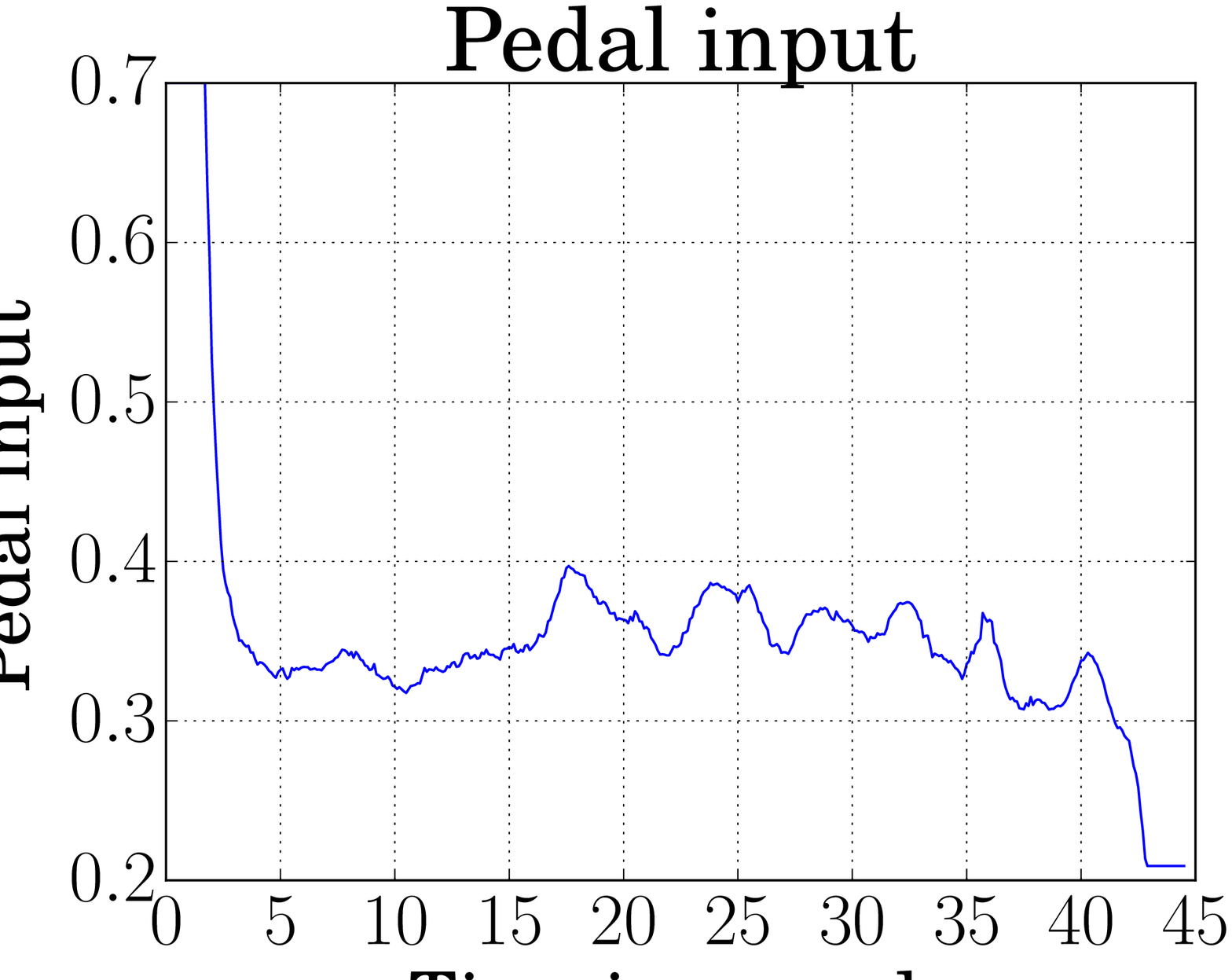}} &
 \subcaptionbox{\label{f4:steer}}{\includegraphics[scale=0.175]{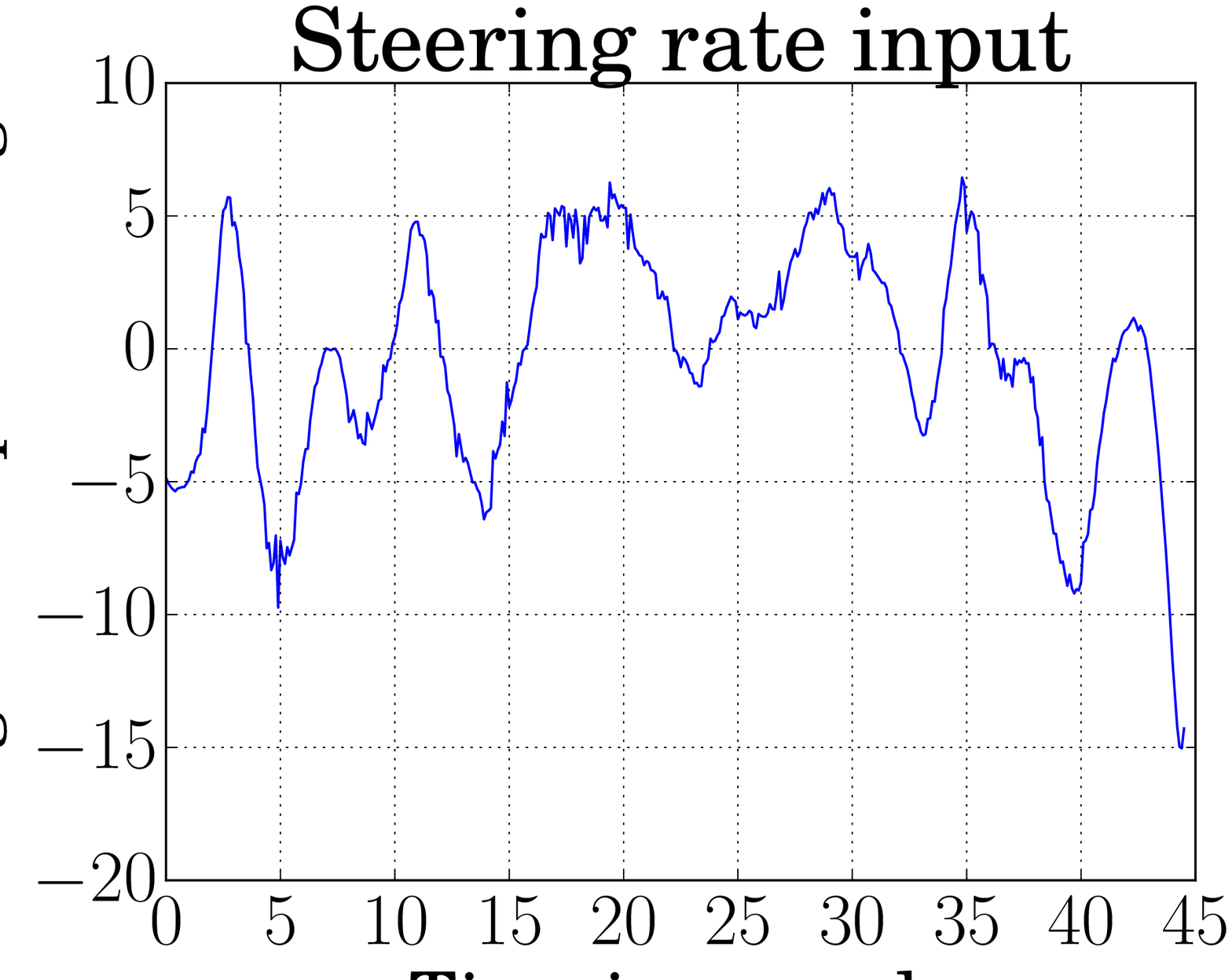}} &
 \subcaptionbox{\label{f4:brake}}{\includegraphics[scale=0.175]{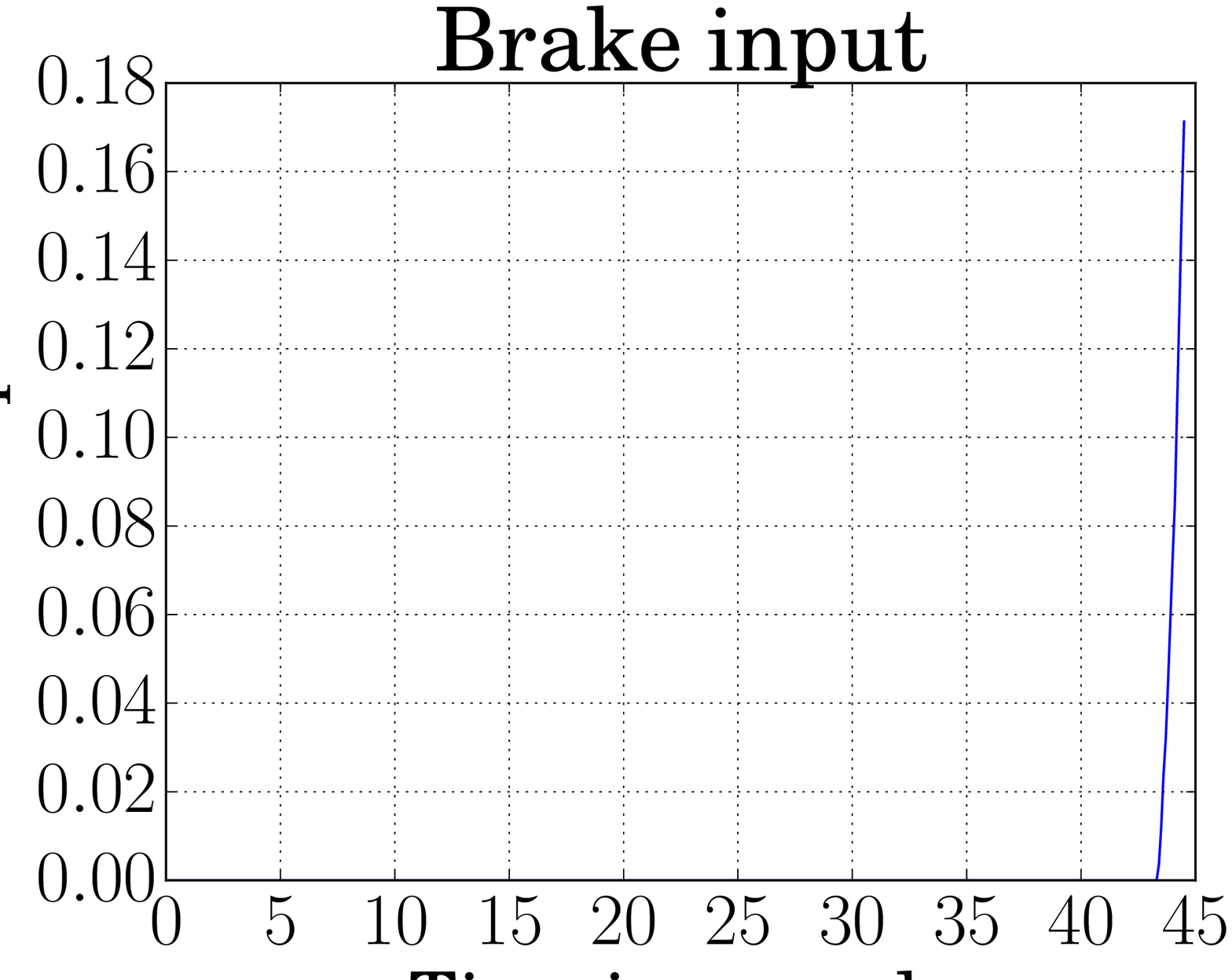}} \\
\end{tabular}
\caption{Polaris GEM e6 Eight trajectory response}
\label{eight}
\end{figure}

\begin{figure}[H]
\onecolumn
\begin{tabular}{ccccc}
\centering
 \subcaptionbox{\label{f5:traj}}{\includegraphics[scale=0.175]{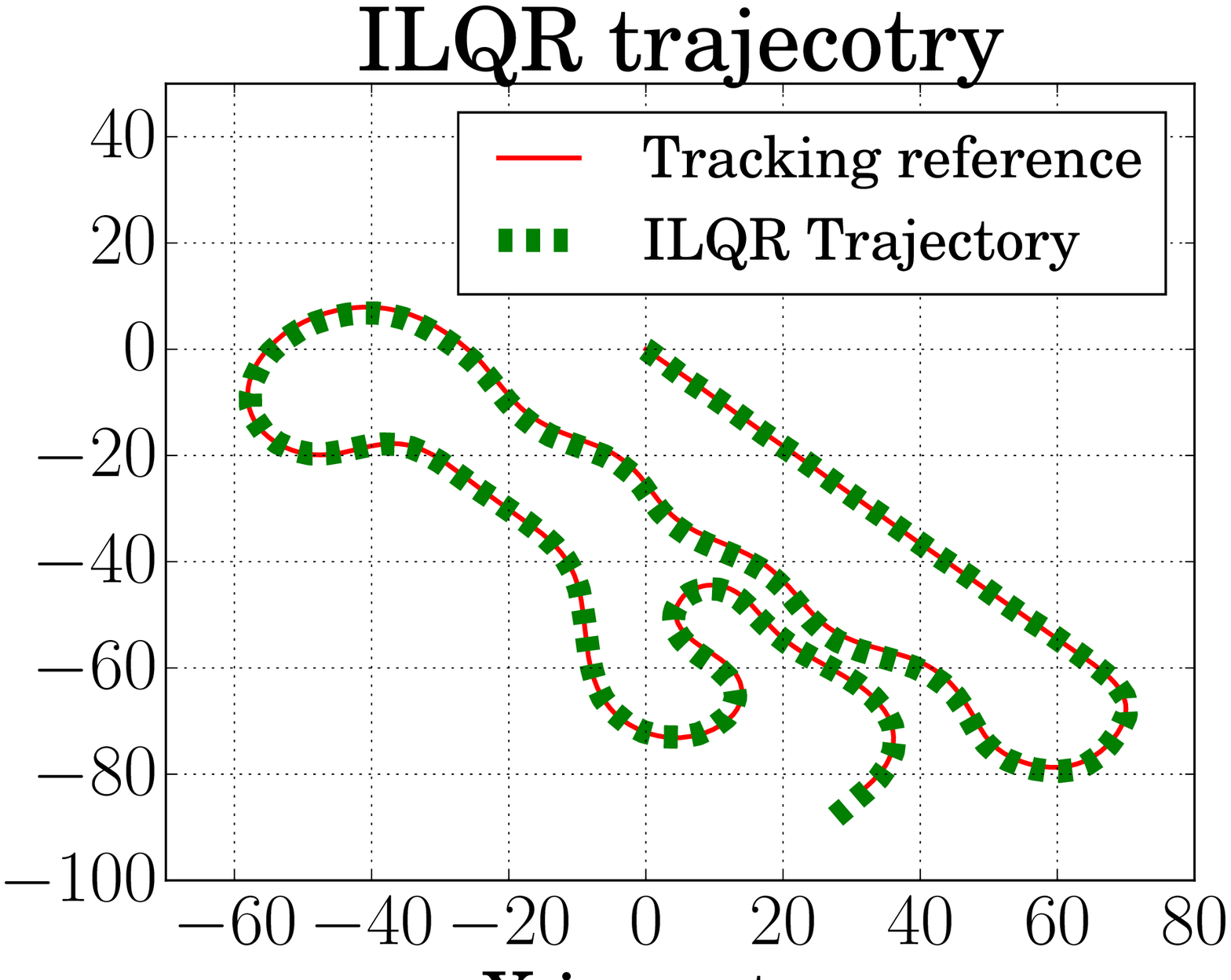}} &
 \subcaptionbox{\label{f5:vel}}{\includegraphics[scale=0.175]{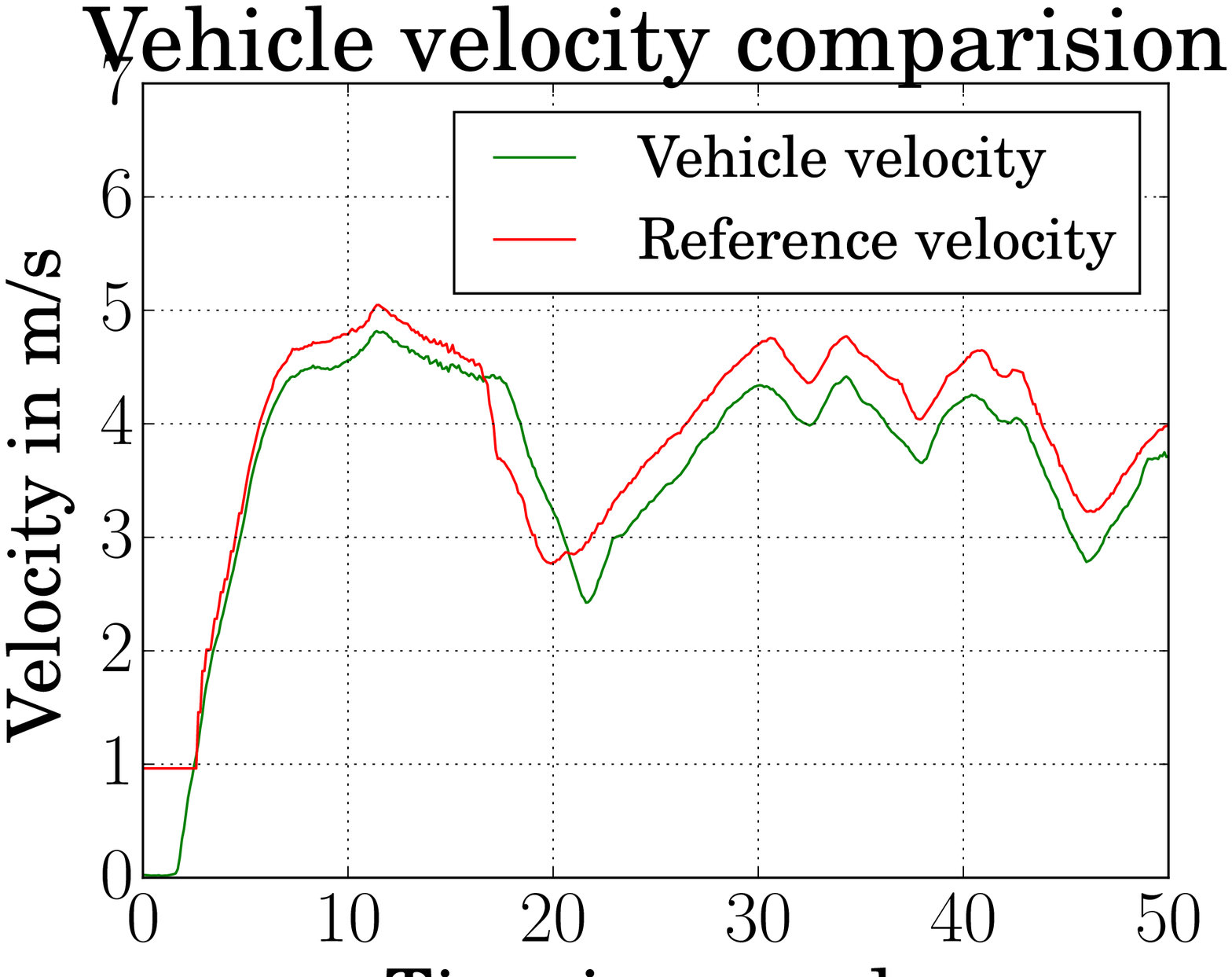}} &
 \subcaptionbox{\label{f5:pedal}}{\includegraphics[scale=0.175]{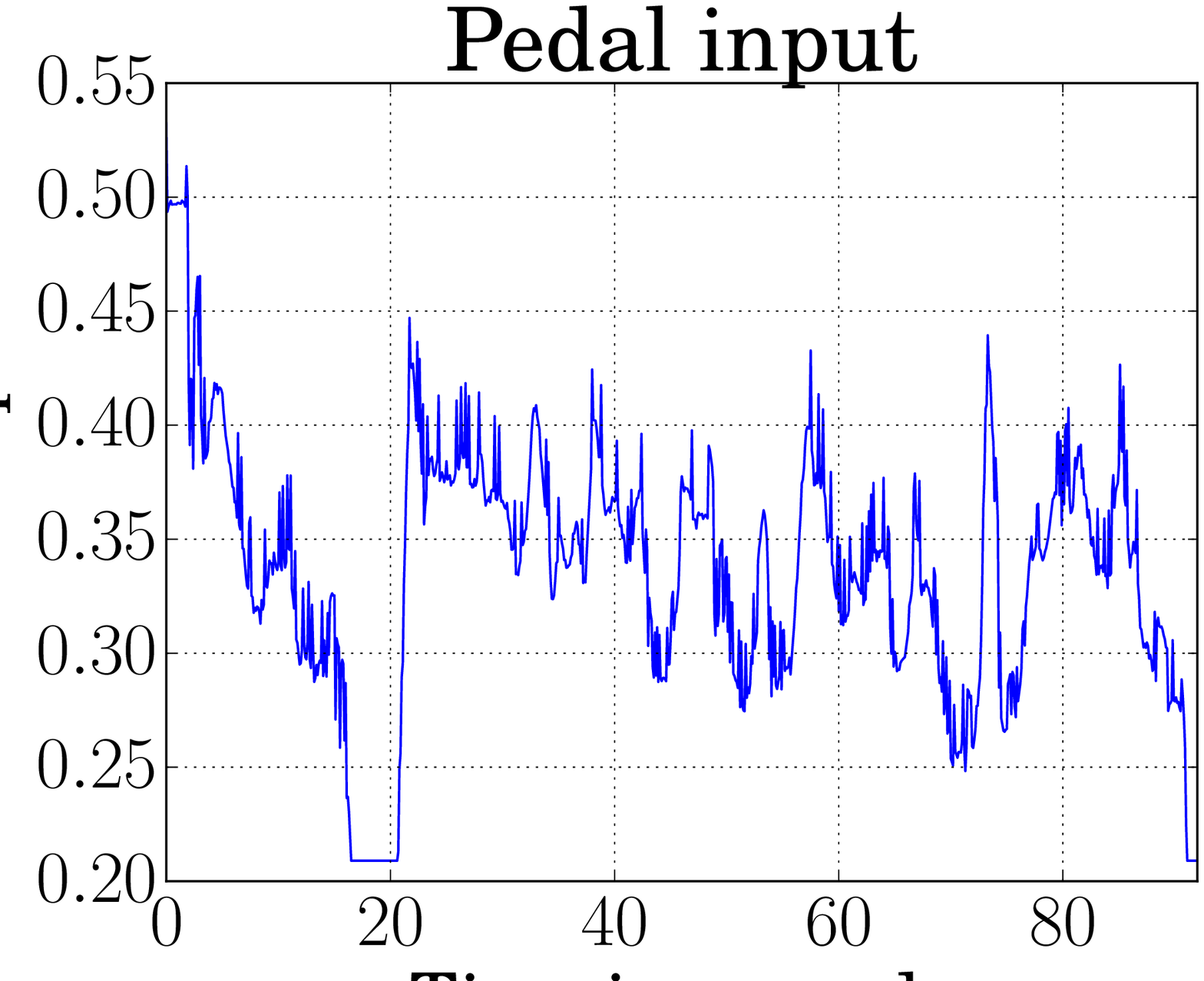}} &
 \subcaptionbox{\label{f5:steer}}{\includegraphics[scale=0.175]{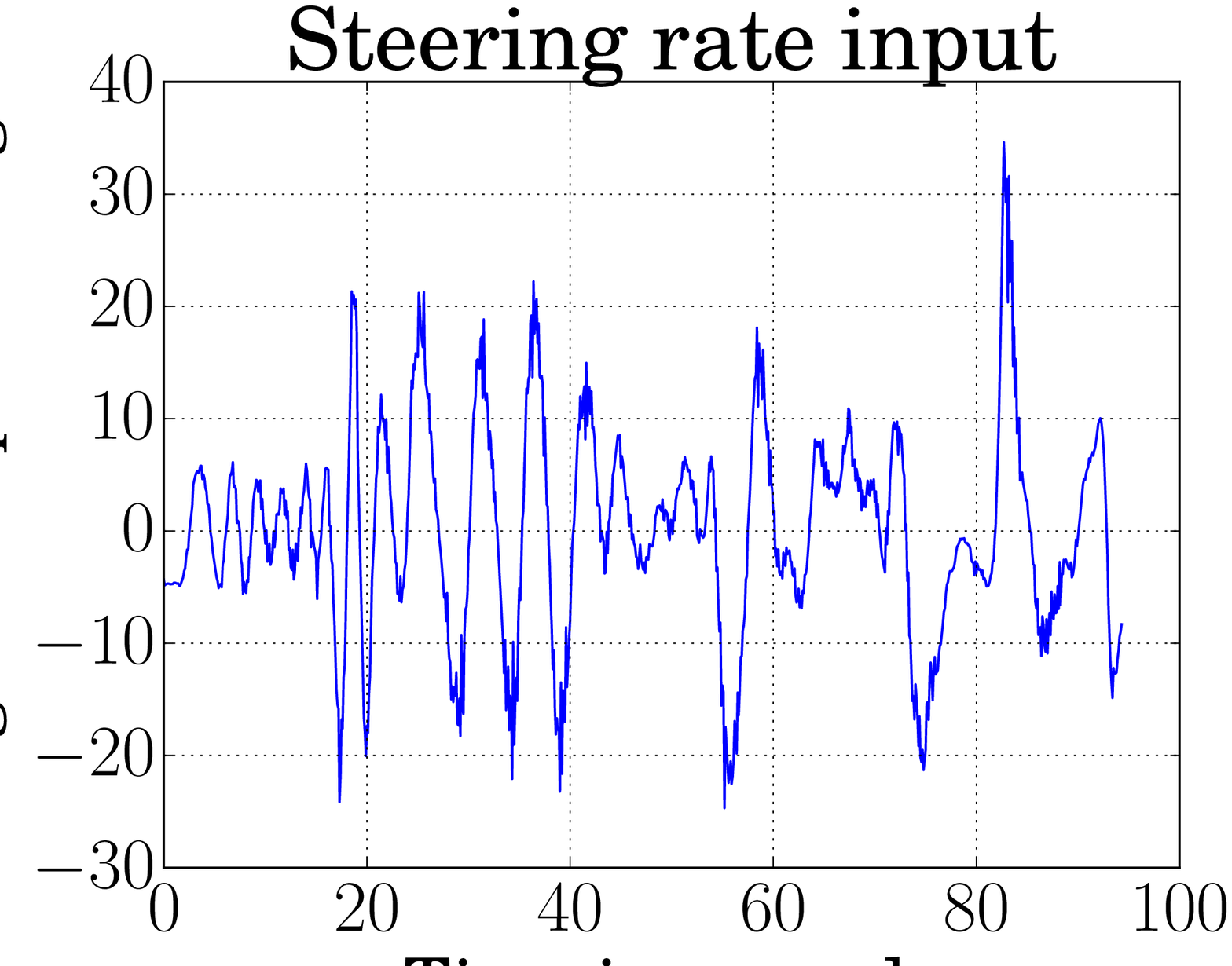}} &
 \subcaptionbox{\label{f5:brake}}{\includegraphics[scale=0.175]{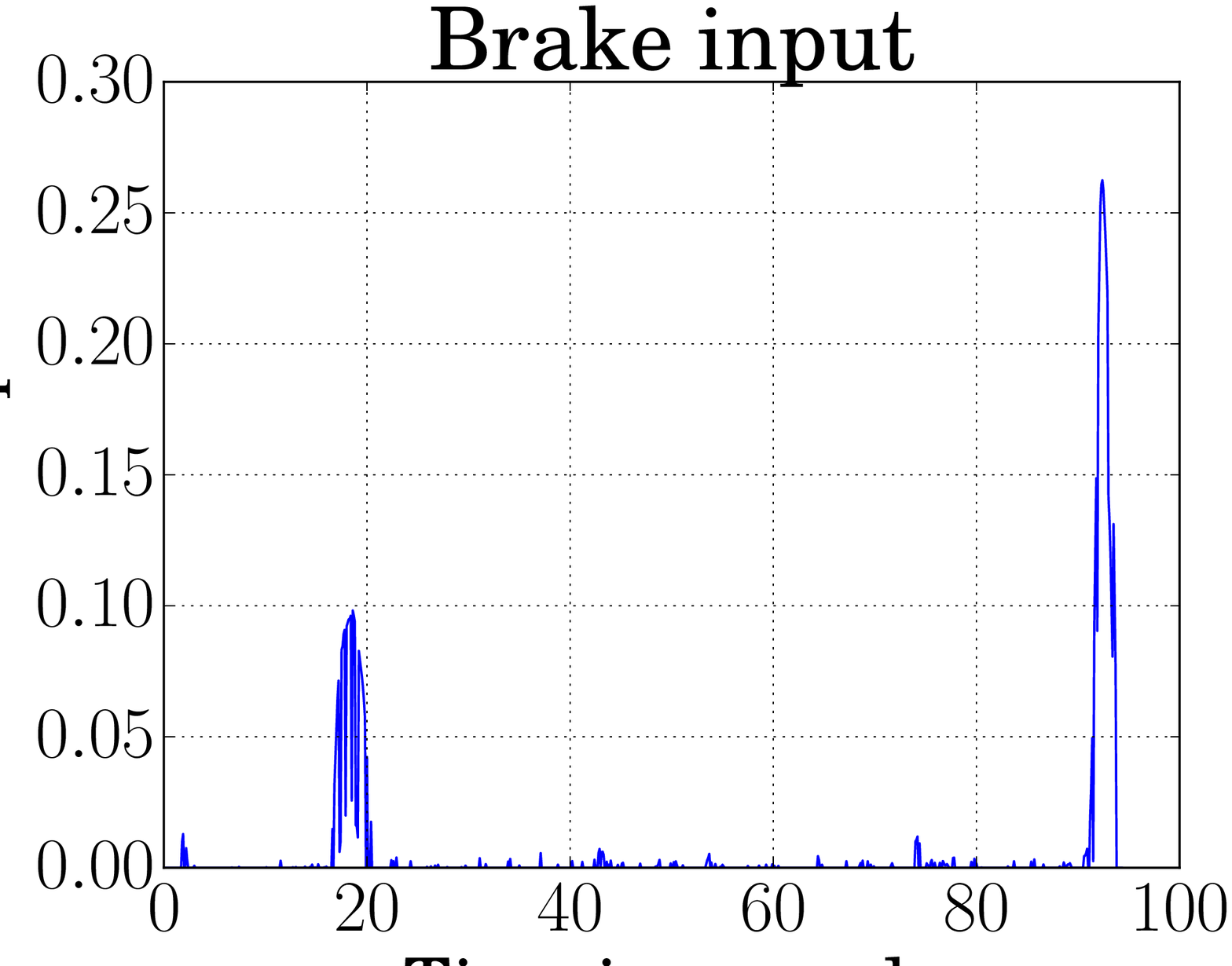}} 
\end{tabular}
\caption{Polaris GEM e6 Combination trajectory response}
\label{comb}
\end{figure}

We evaluate the performance of the trajectory tracking algorithm by using four metrics,
average cross track error(ACE), maximum cross track error(MCE), average velocity error(AVE) and maximum velocity(MVE). For Polaris GEM e6
we calculate these metrics on five types of reference trajectories namely circular track `Fig. \ref{f1:traj}', oval track `Fig. \ref{f2:traj}', snake track `Fig. \ref{f3:traj}', 'Eight' track `Fig. \ref{f4:traj}',
and a combination track  `Fig. \ref{f5:traj}'. We collect the reference trajectories by logging the VectorNav-300 GPS data while driving manually.
We evaluate Warthog's performance on the reference trajectory shown in `Fig. \ref{f6:traj}' 
which involves moving at the speeds of 3m/s-4m/s,
mimicking the kind of trajectories Warthog might be
required to follow in an off-road environment. The warthog has a maximum velocity of 4.5m/s so
we are testing the controller at the limits of what the Warthog can perform.

`Table \ref{errom}' summarizes the results of these experiments on both Warthog and Polaris GEM e6. 
Both vehicles are equipped with a VectorNav-300 GPS for localization which is accurate up to 20-30 cm, considering the GPS accuracy 
the ACEs and MCEs are acceptable for both Polaris GEM e6 and Warthog. The reason for high MVEs is the fact that
at the start vehicle has zero velocity while the initial points in the reference trajectories have 1m/s-2m/s velocities.
The high MCE for 'Eight' track is due to the 0.8m/s reference velocity reported by GPS around 25th second from starting 
time as shown in `Fig. \ref{f4:vel}'. 

`Fig. [4-9](b)' compare commanded velocities and actual vehicle velocities for the reference trajectories.
The control input plots `Fig. [4-8][c-e]' and `Fig. 9[c-d]' 
show that the control inputs satisfy the predefined constraints of $[0,1]$ for pedal and brake, $[-60,60]$ for steering rate(deg/s), $[0,4.5]$ for Warthog linear velocity (m/s) and $[-180, 180]$ for Warthog angular velocity (deg/s).



In this work we demonstrated a model based control methodology for an off-road vehicle as well as an on-road shuttle with varying dynamics, speeds as well as environmental conditions.
In future we
plan to compare this approach with classical geometric and dynamic controllers for wheeled robots in trajectory following context. 
We also plan to 
implement the controller presented in this paper on an eighteen wheeler with a trailer attached. Truck and a trailer has a complex non linear dynamics and is 
a challenging problem what will further test the limits of the controller presented in this work.

\vspace{0.4cm}

\newpage
\setlength{\tabcolsep}{0.01em}
 
 \vspace{-0.8cm}
 
 \vspace{-0.8cm}

  \vspace{-0.8cm}

 \vspace{-0.8cm}

 \vspace{-0.8cm}

\begin{figure}[H]
\onecolumn
\setlength{\tabcolsep}{1em}
\begin{tabular}{cccc}
\centering
 \subcaptionbox{\label{f6:traj}}{\includegraphics[scale=0.175]{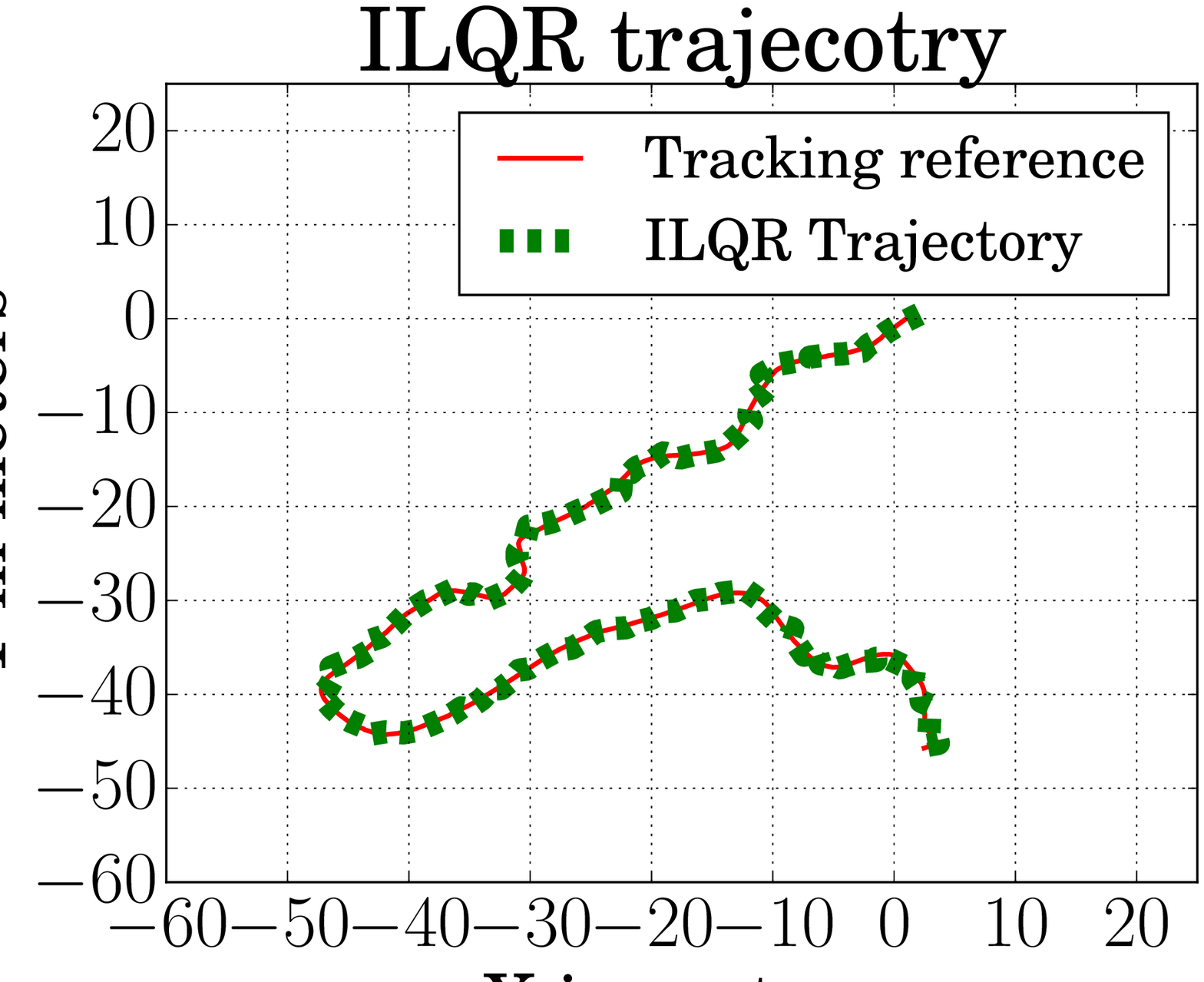}} &
 \subcaptionbox{\label{f6:vel}}{\includegraphics[scale=0.175]{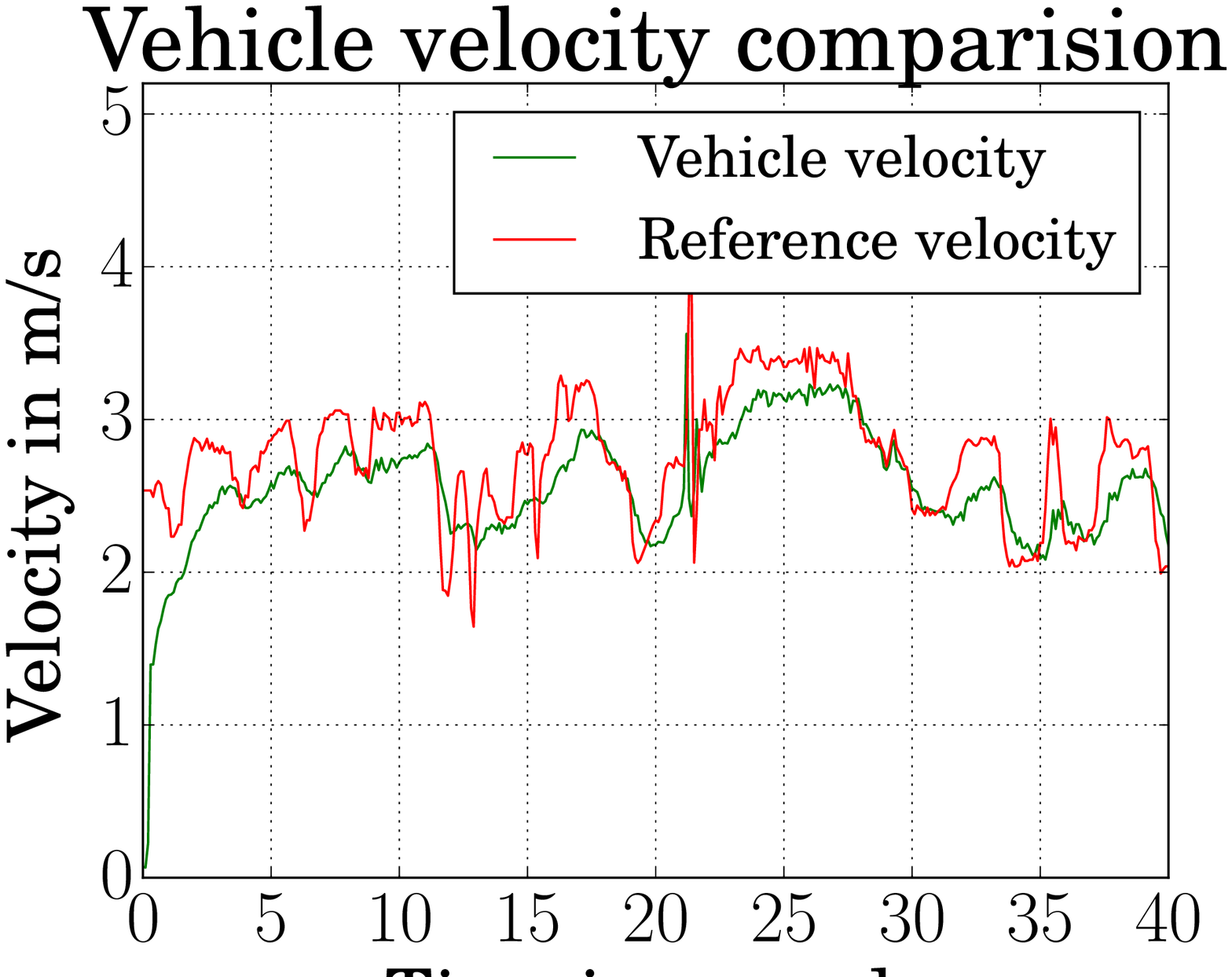}} &
 \subcaptionbox{\label{f6:lin}}{\includegraphics[scale=0.175]{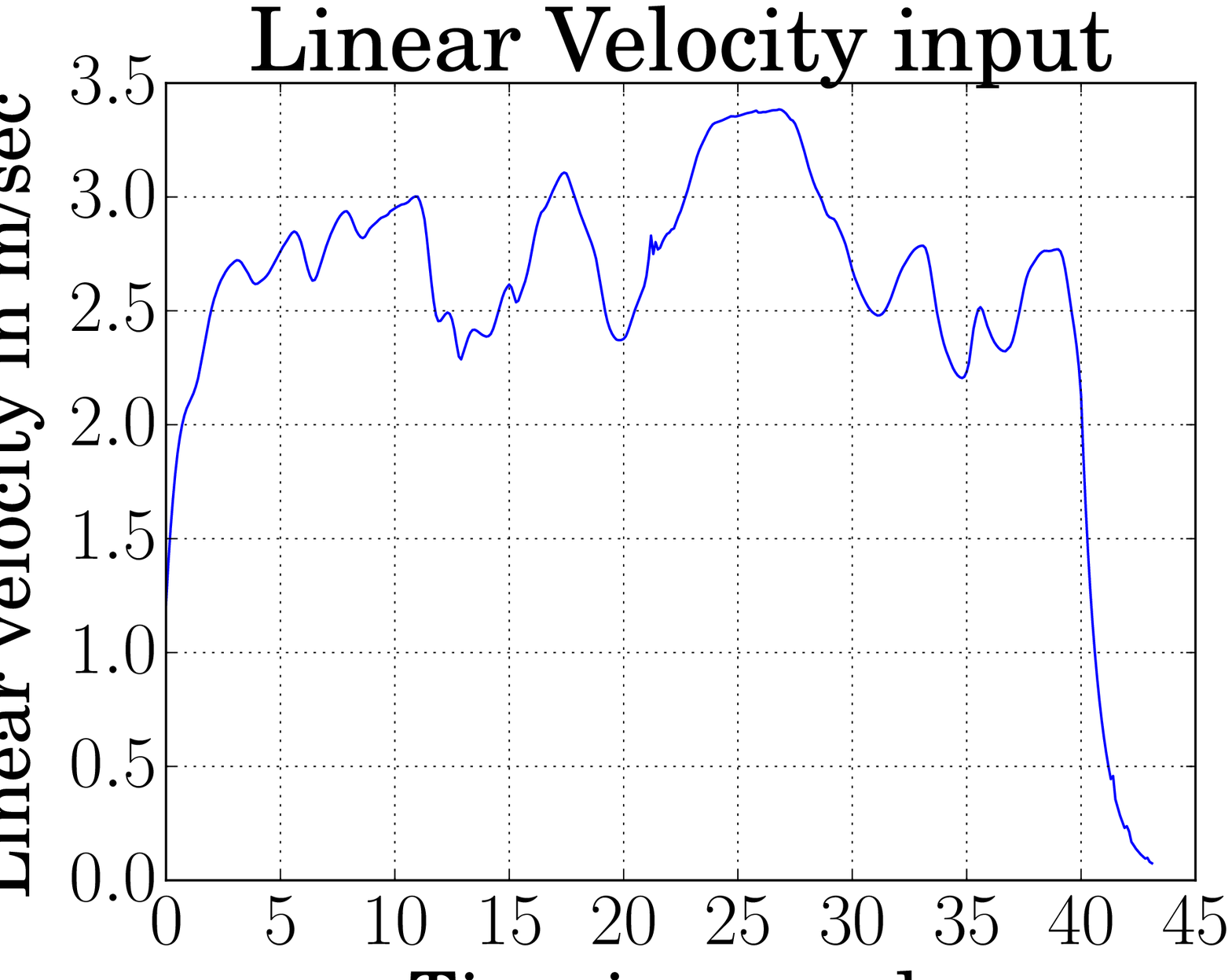}} &
 \subcaptionbox{\label{f6:ang}}{\includegraphics[scale=0.175]{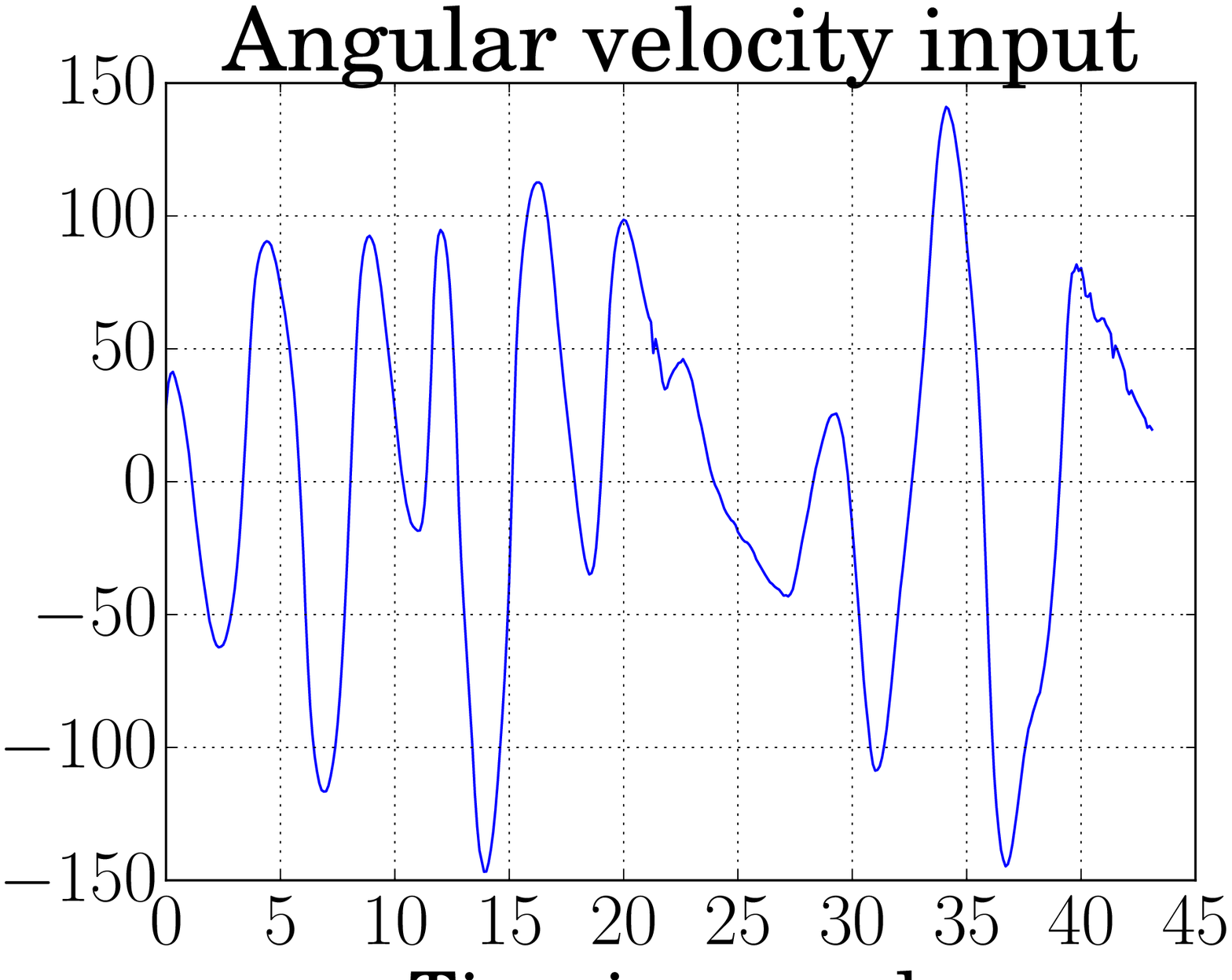}} 
\end{tabular}
\caption{Warthog Combination trajectory response}
\label{warthog}
\end{figure}
 
 \twocolumn 

\addtolength{\textheight}{-12cm}   









\begin{table}
\begin{tabular}{|p{2.9cm}|p{1.4cm}|p{1.4cm}|p{1.4cm}|p{1.4cm}|}
 \hline
 \multicolumn{5}{|c|}{Error Metrics}\\
 \hline
 Reference & ACE & MCE & AVE & MVE \\
 \hline
 GEM Circular & 0.24m & 0.61m & 0.44m/s & 1.73m/s \\
 \hline
 GEM Oval & 0.32m & 0.66m & 0.48m/s & 1.61m/s \\
 \hline
 GEM Snake & 0.40m & 0.72m & 0.46m/s & 1.02m/s \\
 \hline
 GEM Eight & 0.44m & 1.32m & 0.58m/s & 3.88m/s \\
 \hline 
 GEM Combination & 0.43m & 0.89m & 0.43m/s & 2.10m/s \\
 \hline
 Warthog Combination & 0.25m & 0.56m & 0.28m/s & 2.46m/s \\
 \hline 
\end{tabular}
 \caption{Error result on various reference Trajectories.} 
 \label{errom}
 \end{table}
\bibliographystyle{IEEEtran}
\bibliography{IEEEabrv,IEEEexample}

\end{document}